\documentclass{cstr}

\usepackage{amsmath,amsfonts,amssymb,latexsym,bm}
\usepackage{amsthm}
\usepackage{mathrsfs}
\usepackage[pdftex]{graphicx}
\usepackage{subfig}
\usepackage{url}
\usepackage{cite}

\usepackage{booktabs, multirow}

\usepackage{algorithm, algorithmic}
\renewcommand{\algorithmicrequire}{{\bf Input (for worker-side):}}
\renewcommand{\algorithmicensure}{{\bf Output (for worker-side):}}

\title{
Another Use of SMOTE for Interpretable Data Collaboration Analysis
}

\author[1,*]{Akira Imakura}
\author[1]{Masateru Kihira}
\author[1]{Yukihiko Okada}
\author[1]{Tetsuya Sakurai}

\affil[1]{University of Tsukuba, 1-1-1 Tennodai, Ibaraki, Tsukuba 305-8573, Japan}

\email{imakura@cs.tsukuba.ac.jp}

\begin{document}
\maketitle

\begin{abstract}
  Recently, data collaboration (DC) analysis has been developed for privacy-preserving integrated analysis across multiple institutions.
  DC analysis centralizes individually constructed dimensionality-reduced {\it intermediate representations} and realizes integrated analysis via {\it collaboration representations} without sharing the original data.
  To construct the collaboration representations, each institution generates and shares a shareable {\it anchor dataset} and centralizes its intermediate representation.
  Although, random anchor dataset functions well for DC analysis in general, using an anchor dataset whose distribution is close to that of the raw dataset is expected to improve the recognition performance, particularly for the interpretable DC analysis.
  Based on an extension of the synthetic minority over-sampling technique (SMOTE), this study proposes an anchor data construction technique to improve the recognition performance without increasing the risk of data leakage.
  Numerical results demonstrate the efficiency of the proposed SMOTE-based method over the existing anchor data constructions for artificial and real-world datasets.
  Specifically, the proposed method achieves 9 percentage point and 38 percentage point performance improvements regarding accuracy and essential feature selection, respectively, over existing methods for an income dataset.
  The proposed method provides another use of SMOTE not for imbalanced data classifications but for a key technology of privacy-preserving integrated analysis.
\end{abstract}

\section{Introduction}
\subsection{Background and motivation}
There is a growing demand for the integrated analysis of data owned by multiple organizations in a distributed manner \cite{feng2022vertical,ni2022federated,imakura2021interpretable}.
In some real-world applications, such as financial, medical, and manufacturing data analyses, it is difficult to share the original data for analysis because of data confidentiality, and privacy-preserving analysis methods, in which datasets are collaboratively analyzed without sharing the original data.
\par
Motivating examples are interorganizational and intraorganizational data collaborations.
An interorganizational example is about a relationship among companies and banks.
Typically, a company borrows from and transacts with a few banks.
The credit and financial data are distributed among borrowing and transacting banks.
Despite the difficulty of sharing distributed data, creditors and analysts desire to collaborate for credibility and future profitability predictions.
\par
On the other hand, intraorganizational collaborations may be needed.
Universities have their entrance examination, learning processes, and healthcare students' data.
However, the data are distributed and cannot be used for a smarter campus life.
Despite the difficulty of sharing distributed data, students' mentors and counselors desire to collaborate to improve diagnosis and advice.
\par
The federated learning systems \cite{konevcny2016federated} that Google introduced are attracting research attention as typical technologies for this topic.
However, the conventional federated learning requires cross-institutional communication during each iteration \cite{li2019survey,konevcny2016federated,mcmahan2016communication,yang2019federated,ni2022federated,feng2022vertical}.
An integrated analysis of multiple institutions motivated this study.
In this scenario, many cross-institutional communications can be a significant issue in social implementation.
\subsection{Main purpose and contributions}
We focus on {\it data collaboration (DC) analysis}, a non-model share-type federated learning that has recently been developed for supervised learning \cite{imakura2020data,imakura2021collaborative,imakura2021interpretable}, novelty detection \cite{imakura2021collaborative2}, and feature selection \cite{ye2019distributed}.
DC analysis centralizes the dimensionality-reduced {\it intermediate representations}.
The centralized intermediate representations are transformed to incorporable forms called {\it collaboration representations} using a shareable {\it anchor dataset}.
Then, the collaborative representation is analyzed as a single dataset.
Unlike federated learnings, DC analysis does not require iterative computations with cross-institutional communications.
\par
The DC analysis performance strongly depends on the anchor dataset, although random anchor data functions well in general \cite{imakura2020data,imakura2021collaborative,imakura2021collaborative2}.
The use of the anchor dataset, whose distribution is close to that of the raw dataset, is expected to improve the recognition performance of DC analysis.
However, using the anchor dataset such that samples are close to the raw data samples causes data leakage.
The anchor data establishment is essential for both the recognition performance and privacy of DC analysis.
\par
This study specifically focuses on the interpretable DC analysis \cite{imakura2021interpretable} which constructs an interpretable model based on DC framework.
Then, we propose an anchor data construction technique based on an extension of the synthetic minority over-sampling technique (SMOTE) \cite{chawla2002smote}, which is a data augmentation method for classification of imbalanced datasets, to improve the recognition performance without increasing the risk of data leakage.
The main contributions of this study are
\begin{itemize}
  \item We propose an anchor data construction technique based on an extension of SMOTE for DC analysis which is a recent non-model share-type federated learning for privacy-preserving integrated analysis.
  \item The proposed SMOTE-based method improves the recognition performance of the interpretable DC analysis without increasing the risk of data leakage.
  \item Numerical results demonstrate the efficiency of the proposed SMOTE-based method over the existing anchor data constructions owing the contribution of the extension of SMOTE.
  \item The proposed method provides another use of SMOTE not for imbalanced data classifications but for a key technology of privacy-preserving integrated analysis.
\end{itemize}
\section{Related works}
\subsection{Federated Learning}
Recently, federated learning systems have been developed for distributed data analysis and privacy preservation.
The concept of federated learning was first proposed by Google \cite{konevcny2016federated} typically for Android phone model updates \cite{mcmahan2016communication}.
Federated learning is primarily based on (deep) neural networks and updates iteratively the model \cite{li2019survey,konevcny2016federated,mcmahan2016communication,yang2019federated,ni2022federated,feng2022vertical}.
\par
Federated stochastic gradient descent (FedSGD) and federated averaging (FedAvg) are standard techniques for updating the model \cite{mcmahan2016communication}. 
FedSGD is a direct extension of the stochastic gradient descent method.
During each iteration of the gradient descent method, each party locally computes a gradient from the shared model using the local dataset before sending it to the server.
The shared gradients are averaged and used to update the model.
In FedAvg, each party performs multiple batch updates using the local dataset and sends the updated model to the server.
Then, the shared models are update via averaging.
These federated learning also including more recent methods such as FedProx \cite{li2020federated} and FedCodl \cite{ni2022federated}, requires cross-institutional communication in each iteration.
\par
For more details, we refer to \cite{li2019survey,yang2019federated} and references therein.
\subsection{Interpretable data collaboration (DC) analysis}
Here, we describe DC analysis for analyzing the following horizontally and vertically partitioned data:
\begin{equation*}
  X = \left[
    \begin{array}{cccc}
      X_{1,1} & X_{1,2} & \cdots & X_{1,d} \\
      X_{2,1} & X_{2,2} & \cdots & X_{2,d} \\
      \vdots & \vdots & \ddots & \vdots \\
      X_{c,1} & X_{c,2} & \cdots & X_{c,d}
    \end{array}
  \right] \in \mathbb{R}^{n \times m}, \quad
  Y = \left[
    \begin{array}{c}
      Y_{1} \\
      Y_{2} \\
      \vdots \\
      Y_{c} 
    \end{array}
  \right].
\end{equation*}
Note that DC analysis is applicable to more complicated data distributions \cite{mizoguchi2022application, imakura2021collaborative}.
\par
DC analysis operates in two roles: {\it worker} and {\it master}.
Workers have the private dataset $X_{i,j} \in \mathbb{R}^{n_i \times m_j}$ ($n = \sum_{i=1}^c n_i$ and $m = \sum_{j=1}^d m_j$) and corresponding ground truth $Y_i$, which must be analyzed without sharing $X_{i,j}$.
\par
First, all workers generate the same anchor dataset $X^{\rm anc} = [X_{:,1}^{\rm anc}, X_{:,2}^{\rm anc}, \dots$, $X_{:,d}^{\rm anc}] \in \mathbb{R}^{r \times m}$, which is shareable data consisting of public or dummy data randomly constructed.
Then, using a dimensionality reduction function $f_{i,j}$, each worker constructs dimensionality-reduced intermediate representations 
\begin{equation*}
  \widetilde{X}_{i,j} = f_{i,j}(X_{i,j}) \in \mathbb{R}^{n_i \times \widetilde{m}_{i,j}}, \quad
  \widetilde{X}_{i,j}^{\rm anc} = f_{i,j}(X_{:,j}^{\rm anc}) \in \mathbb{R}^{r \times \widetilde{m}_{i,j}}
\end{equation*}
where $\widetilde{m}_{i,j} < m_j$, and centralizes them to the master.
A typical setting for dimensionality reduction function is non-supervised dimensionality reduction methods, such as principal component analysis (PCA) \cite{jolliffe1986principal}, locality preserving projection (LPP) \cite{he2004locality}, and nonnegative matrix factorization (NMF) \cite{lee2000algorithms} and supervised dimensionality reduction methods, such as linear discriminant analysis (LDA) \cite{fisher1936use}, local Fisher discriminant analysis (LFDA) \cite{sugiyama2007dimensionality}, Locality adaptive discriminant analysis (LADA) \cite{li2017locality}, and complex moment-based supervised eigenmap (CMSE) \cite{imakura2019complex}.
\par
On the master-side, the following collaboration representation
\begin{equation*}
  \widehat{X}_{i,j} = g_{i}(\widetilde{X}_i) \in \mathbb{R}^{n_i \times \widehat{m}}, \quad
  \widetilde{X}_i = [\widetilde{X}_{i,1}, \widetilde{X}_{i,2}, \dots, \widetilde{X}_{i,d}] \in \mathbb{R}^{n_i \times \widetilde{m}_{i}}
\end{equation*}
is set such that collaboration representations of the anchor data are approximately the same, where $\widetilde{m}_i = \sum_{j=1}^d \widetilde{m}_{i,j}$, practically by solving a minimal perturbation problem.
The collaboration representations are then analyzed as a single dataset.
We obtain the prediction result of the anchor data by applying the obtained model to the anchor data's collaboration representation.
Sharing the prediction result of the anchor data with the workers, interpretable model are constructed on the worker-side using the anchor data and its prediction result.
\par
The algorithm of the interpretable DC analysis is shown in Algorithm~\ref{alg:IDC}.
For details, please refer to \cite{imakura2021interpretable}.
\begin{algorithm}[!t]
\caption{Interpretable data collaboration analysis}
\label{alg:IDC}
\begin{algorithmic}
  \REQUIRE $X_{i,j} \in \mathbb{R}^{n_i \times m_{j}}$ and $Y_i$ individually
  \ENSURE Interpretable models $t_i$ $(i = 1, 2, \dots, c)$
  \STATE
  \STATE
  \begin{tabular}{rcll}
    & \multicolumn{2}{c}{ {\it Worker-side} $(i,j)$} & \\ \cmidrule{2-3}
    1:  & \multicolumn{2}{l}{Generate $X^{\rm anc}$ and share with all workers} & \\
    2:  & \multicolumn{2}{l}{Generate $f_{i,j}$} & \\
    3:  & \multicolumn{2}{l}{Compute $\widetilde{X}_{i,j} = f_{i,j}(X_{i,j})$ and $\widetilde{X}^{\rm anc}_{i,j} = f_{i,j}(X_{:,j}^{\rm anc})$} & \\
    4:  & \multicolumn{2}{l}{Share $\widetilde{X}_{i,j}, \widetilde{X}_{i,j}^{\rm anc}$, and $Y_i$ to master} & \\
    \\
    &   & \multicolumn{2}{c}{ {\it Master-side}}  \\ \cmidrule{3-4}
    5:  & \qquad $\searrow$ & \multicolumn{2}{l}{Obtain $\widetilde{X}_{i,j}, \widetilde{X}_{i,j}^{\rm anc}$, and $Y_i$ for all $i$ and $j$}  \\
    6:  & & \multicolumn{2}{l}{Set $\widetilde{X}_i, \widetilde{X}_i^{\rm anc}$, and $Y$} \\
    7:  & & \multicolumn{2}{l}{Compute $G_i$ from $\widetilde{X}_{i}^{\rm anc}$ for all $i$} \\
    8:  & & \multicolumn{2}{l}{Compute $\widehat{X}_{i} = \widetilde{X}_{i} G_i$ for all $i$, and set $\widehat{X}$} \\
    9:  & & \multicolumn{2}{l}{Analyze $\widehat{X}$ and $Y$ to obtain $h(\widehat{X}) \approx Y$} \\
    8:  & & \multicolumn{2}{l}{Compute $Y^{\rm anc}_i = h(\widetilde{X}_{i}^{\rm anc}G_i)$ for all $i$} \\
    11: & \qquad $\swarrow$ & \multicolumn{2}{l}{Return $Y_i^{\rm anc}$ to each worker} \\
    \\
    & \multicolumn{2}{c}{ {\it Worker-side} $(i,j)$} & \\ \cmidrule{2-3}
    12: & \multicolumn{2}{l}{Obtain $Y_i^{\rm anc}$} \\
    13: & \multicolumn{2}{l}{Analyze $X^{\rm anc}$ and $Y_i^{\rm anc}$ to obtain $t_i(X^{\rm anc}) \approx Y_i^{\rm anc}$} \\
  \end{tabular}
\end{algorithmic}
\end{algorithm}
\subsection{SMOTE}
Classification performance is negatively impacted by imbalanced data for classification problems when the number of samples for each label in the training dataset differ significantly.
In such scenario, under-sampling techniques, which remove majority data with many samples, and over-sampling techniques, which generate minority data with few samples, are used.
Particularly, when the imbalance ratio is high, under-sampling methods must remove many samples, resulting in the deterioration of classification performance.
\par
SMOTE is the most typical and pioneering work of over-sampling technique \cite{chawla2002smote}.
SMOTE generates a new dataset using a randomized interpolation with $k$ nearest neighbors.
Let ${\bm x}_i$ be a minority class sample and $\check{\bm x}_i^{(1)}, \check{\bm x}_i^{(2)}, \dots, \check{\bm x}_i^{(\ell)}$ be $\ell$ samples randomly selected from $k$ nearest neighbors with the same label, of which $\ell \leq k$, where $k$ is generally set to a small value (five by default).
Then, the new data is generated by
\begin{equation*}
  {\bm x}_i^{(j)} = {\bm x}_i + c_{i,j} (\check{\bm x}_i^{(j)} - {\bm x}_i), \quad
  j = 1, 2, \dots, \ell,
\end{equation*}
where $c_{i,j}$ denotes a coefficient of the randomized interpolation, which is randomly set in $[0,1]$.
\par
Because SMOTE uniformly generates the new data from the minority samples, it increases the number of samples that are not necessarily important for classification.
Therefore, improvement methods that intensively generate samples near the classification boundary, such as ADASYN \cite{he2008adasyn}, borderline SMOTE \cite{han2005borderline}, and safe-level SMOTE \cite{bunkhumpornpat2009safe}, have been proposed.
\section{Another use of SMOTE for interpretable data collaboration (DC) analysis}
Let $n$ and $m$ be the number of samples and the dimensionality of the data, respectively.
Additionally, let $X = [{\bm x}_1, {\bm x}_2, \dots, {\bm x}_n]^{\rm T} \in \mathbb{R}^{n \times m}$ and $Y = [{\bm y}_1, {\bm y}_2, \dots, {\bm y}_n] \in \mathbb{R}^{n \times \ell}$ be a data matrix and the corresponding ground truth, respectively.
In this study, for privacy-preserving analysis of multiple parties, we consider horizontal and vertical data partitioning, where data samples and features are partitioned into $c$ and $d$ parties, respectively, as follows:
\begin{align}
  X = \left[
    \begin{array}{cccc}
      X_{1,1} & X_{1,2} & \cdots & X_{1,d} \\
      X_{2,1} & X_{2,2} & \cdots & X_{2,d} \\
      \vdots & \vdots & \ddots & \vdots \\
      X_{c,1} & X_{c,2} & \cdots & X_{c,d}
    \end{array}
  \right], \quad
  Y = \left[ 
    \begin{array}{c}
      Y_1 \\ Y_2 \\ \vdots \\ Y_c
    \end{array}
  \right].
  \label{eq:data}
\end{align}
Additionally, we assume that we have a small amount of shareable public data $X^{\rm pub} \in \mathbb{R}^{p \times m}$ with $p \ll n$.
Note that there are no ground truth data for $X^{\rm pub}$.
\subsection{Existing anchor data construction}
In most of the existing studies, \cite{imakura2020data,imakura2021collaborative,imakura2021collaborative2}, anchor data $X^{\rm anc} \in \mathbb{R}^{r \times m}$ were constructed as a random matrix in the range of the corresponding features as follows:
\begin{equation*}
  X^{\rm anc} = [x_{i,j}^{\rm anc}]_{1\leq i \leq r, 1 \leq j \leq m}, \quad
  x_{i,j}^{\rm anc} \sim \mathcal{U}({x}_j^{\min}, {x}_j^{\max}),
\end{equation*}
where ${x}_j^{\min}$ and ${x}_j^{\max}$ denote the minimum and maximum values of the $j$-th feature in the original data $X$, respectively, and $\mathcal{U}(a,b)$ denotes the uniform distribution on range $[a,b]$.
Hereafter, we will refer to this method as a random anchor data construction method.
\par
In \cite{imakura2021interpretable} for interpretable DC collaboration analysis, a low-rank approximation method was used to construct the anchor data closer to the raw data.
In each worker, the local anchor data $X_{i,j}^{\rm approx}$ are constructed using a low-rank approximation of $X_{i,j}$ with random perturbation as follows:
\begin{equation*}
  X_{i,j}^{\rm approx} = X_{i,j}^{\rm TSVD} + \delta E,
\end{equation*}
where $X_{i,j}^{\rm TSVD} \in \mathbb{R}^{n_i \times m_j}$ denotes a low-rank approximation based on the truncated SVD of $X_{i,j}$, $\delta$ denotes a perturbation parameter, and $E \in \mathbb{R}^{n_i \times m_i}$ denotes a random matrix.
By sharing $X_{i,j}^{\rm approx}$ with all users, $n$ samples of anchor data are generated, as follows:
\begin{align*}
  X^{\rm approx} 
  &= [X^{\rm approx}_{:,1}, X^{\rm approx}_{:,2}, \dots, X^{\rm approx}_{:,d}] \\
  &= \left[ 
    \begin{array}{cccc}
      X^{\rm approx}_{1,1} & X^{\rm approx}_{1,2} & \cdots & X^{\rm approx}_{1,d} \\
      X^{\rm approx}_{2,1} & X^{\rm approx}_{2,2} & \cdots & X^{\rm approx}_{2,d} \\
      \vdots & \vdots & \ddots & \vdots \\
      X^{\rm approx}_{c,1} & X^{\rm approx}_{c,2} & \cdots & X^{\rm approx}_{c,d}
    \end{array}
  \right] \in \mathbb{R}^{n \times m}.
\end{align*}
Next, to generate $r$ samples of anchor data $X^{\rm anc}$, we apply an augmentation technique using a linear combination if $r > n$, or we select $r$ samples randomly otherwise (that is, $r \leq n$).
Hereafter, we will refer to this method as a TSVD-based anchor data construction method.
\subsection{Proposal for a SMOTE-based anchor data construction}
In the TSVD-based anchor data construction method, the constructed anchor data $X^{\rm anc}$ is expected to be closer to the raw data $X$ by increasing the rank of $X_{i,j}^{\rm TSVD}$ and decreasing $\delta$, which improves the recognition performance of DC analysis.
However, the data leakage risk increases.
To improve the recognition performance without increasing the data leakage risk, we propose an anchor data construction technique based on an extension of SMOTE.
\par
In each local party, we generate an anchor dataset $X^{\rm anc}$ using a SMOTE-based method from $X^{\rm pub}$ using the same random numbers.
To generate $X^{\rm anc}$ that mimics the distribution of the raw data from $X^{\rm pub}$ with small samples ($p \ll n$), we extend SMOTE to change the range of parameter $c$ as $[0,\alpha]$ to allow even extrapolation.
Note that the conventional SMOTE considers only interpolation.
Additionally, a larger value is taken for the number of neighbors $k$, which is generally set to a small value in conventional SMOTE (five by default).
These broaden the data distribution.
The contribution of these extensions of SMOTE for DC analysis will be evaluated in Section 4.4.
\par
Note that conventional SMOTE is generally used for imbalance data.
In contrast, this study performs over-sampling on all $X^{\rm pub}$ data to construct the anchor dataset that mimics the raw data to improve the recognition performance of DC analysis.
For this reason, the classical SMOTE is used instead of its improvements, which heavily oversamples on the classification boundary, as typified by ANASYN.
\par
The algorithm of the proposed SMOTE-based anchor data construction is summarized in Algorithm~\ref{alg:proposed}.
\par
Here, we analyze the variance of the new samples.
Let ${\bm x}_1$ and ${\bm x}_2$ be the original samples that are independently selected ($k$ of $k$ nearest neighbors are set as $n$) so that it has zero covariance, ${\rm cov}({\bm x}_1,{\bm x}_2) = 0$.
Here, we assume that the expected values of the samples are zero, $E[{\bm x}_1] = E[{\bm x}_2] = 0$.
Additionally, let $c$ be a uniform random number in $[0,\alpha]$ $(\alpha > 0)$.
\par
Then, the variance of the new sample $V[{\bm x}'] = V[{\bm x}_1 + c ( {\bm x}_2 - {\bm x}_1)]$ can be written as follows:
\begin{align*}
  &V[ {\bm x}_1 + c ( {\bm x}_2 - {\bm x}_1)]  \\
  &\quad = V[ (1-c) {\bm x}_1] + V[c {\bm x}_2] \\
  &\quad = V[ (1-c)] V[ {\bm x}_1] + E[1-c]^2 V[{\bm x}_1] + V[c] V[ {\bm x}_2] + E[c]^2 V[{\bm x}_2] \\
  &\quad = \frac{\alpha^2}{12} V[ {\bm x}_1] + \left(1-\frac{\alpha}{2}\right)^2 V[{\bm x}_1] + \frac{\alpha^2}{12} V[ {\bm x}_2] + \left(\frac{\alpha}{2}\right)^2 V[{\bm x}_2] \\
  &\quad = \left( \frac{2}{3} \alpha^2 - \alpha + 1 \right) V[{\bm x}_1],
\end{align*}
where we used $V[c] = \alpha^2/12$ and $E[c] = \alpha/2$.
Thus, we have $V[{\bm x}'] = V[{\bm x}_1]$ when $\alpha =1.5$, $V[{\bm x}'] < V[{\bm x}_1]$ when $\alpha < 1.5$, and $V[{\bm x}'] > V[{\bm x}_1]$ when $\alpha > 1.5$.
\begin{algorithm}[!t]
\caption{A SMOTE-based anchor data construction}
\label{alg:proposed}
\begin{algorithmic}[1]
  \REQUIRE $X^{\rm pub} \in \mathbb{R}^{p \times m}$, the number of anchor data $r$ and parameters $\alpha$ and $k$
  \ENSURE Anchor data $X^{\rm anc} \in \mathbb{R}^{r \times m}$
  \STATE Normalize $X^{\rm pub}$
  \FOR{$i = 1, 2, \dots, p$}
  \STATE Set $k$ nearest neighbors of ${\bm x}_i^{\rm pub}$
  \STATE Randomly select $r/p$ samples $\check{\bm x}_i^{(j)}$ $(j = 1, 2, \dots, r/p)$ from $k$ nearest neighbors
  \FOR{$j = 1, 2, \dots, r/p$}
  \STATE Randomly set $c_{i,j} \in [0, \alpha]$ and compute ${\bm x}_i^{(j)} = {\bm x}_i + c_{i,j} (\check{\bm x}_i^{(j)} - {\bm x}_i)$.
  \ENDFOR
  \ENDFOR
  \STATE Set $X^{\rm anc}$ as all generated vectors ${\bm x}_i^{(j)}$
  \STATE Denormalize $X^{\rm anc}$
\end{algorithmic}
\end{algorithm}
\section{Numerical experiments}
This section evaluates the efficiency of the proposed SMOTE-based anchor data construction method (Algorithm~\ref{alg:proposed}) for the interpretable DC analysis ({\bf DC(SMOTE)}) and compares it with existing anchor data construction techniques: using a random matrix ({\bf DC(rand)}) and a TSVD ({\bf DC(TSVD)}) introduced in Section 3.1.
We also evaluate a scenario in which the raw data is used for the anchor data ({\bf DC(raw)}).
We compared the interpretable DC analysis with the centralized analysis that shares the raw datasets ({\bf Centralized}) and the local analysis that only uses local dataset $X_{i,j}$ ({\bf Local}) to assess prediction accuracy.
Note that {\bf DC(raw)} and {\bf Centralized} are considered ideal cases because the raw data cannot be shared in our target situation.
\subsection{General settings}
We used PCA for dimensionality reduction method on worker-side (Step 2 in Algorithm~\ref{alg:IDC}) for the interpretable DC analysis.
We set $\widehat{m} = \widetilde{m}_i$.
We set $p = 100$ for the number of public data $X^{\rm pub}$.
We also set the ground truth $Y$ as a binary matrix whose $(i,j)$-th entry is 1 if the training data ${\bm x}_i$ are in class $j$ and $0$ otherwise.
This type of ground truth has been applied to various classification algorithms, including ridge regression and deep neural networks \cite{bishop2006pattern}.
\par
Here, we evaluate the efficiency of the anchor data construction methods regarding data confidentiality and recognition performance.
For data confidentiality, we evaluated the similarity between the anchor data and the raw data by
\begin{itemize}
  \item Earth mover's distance (EMD) \\
    \begin{equation*}
      {\rm EMD}(X,X^{\rm anc}) = \min_{f_{i,j}} \sum_{i,j} f_{i,j} \| {\bm x}_i - {\bm x}_j^{\rm anc} \|_2,
    \end{equation*}
    where $f_{i,j} = 0$ or $1$, $\sum_{i=1}^n f_{i,j} = 1$ and $\sum_{j=1}^r f_{i,j} = 1$.
\item Average minimum distance from the raw data (AMD(raw)) \\
    \begin{equation*}
      {\rm AMD(raw)} = {\rm AMD}(X,X^{\rm anc}) = \frac{1}{n} \sum_{i=1}^n \min_{1\leq j \leq r} \| {\bm x}_i - {\bm x}_j^{\rm anc} \|_2.
    \end{equation*}
  \item Average minimum distance from the anchor data (AMD(anc)) \\
    \begin{equation*}
      {\rm AMD(anc)} = {\rm AMD}(X^{\rm anc},X) = \frac{1}{r} \sum_{i=1}^r \min_{1\leq j \leq n} \| {\bm x}_i^{\rm anc} - {\bm x}_j \|_2.
    \end{equation*}
\end{itemize}
For recognition performance, we evaluated the constructed interpretable model by
\begin{itemize}
  \item Accuracy (ACC) of prediction result \\
  ACC is the ratio of correct predictions, defined as
\begin{equation*}
  {\rm ACC}(Y^{\rm GT}, Y^{\rm Pred}) = \frac{ \mbox{number of correct predictions} }{ \mbox{ number of test samples} },
\end{equation*}
where $Y^{\rm GT}$ and $Y^{\rm Pred}$ denote the ground truth and prediction result for the test data, respectively.
  \item Normalized mutual information (NMI) \\
    NMI is the mutual information (MI) score normalized to produce results between 0 (no mutual information) and 1 (perfect correlation), defined as
    \begin{equation*}
      {\rm NMI}(Y^{\rm Pred}, Y^{\rm GT}) = \frac{ I(Y^{\rm Pred}; Y^{\rm GT}) }{ \sqrt{H(Y^{\rm Pred}) H(Y^{\rm GT})} },
    \end{equation*}
    where $Y^{\rm GT}$ and $Y^{\rm pred}$ are the ground truth and prediction result, respectively, and $I(Y^{\rm Pred}; Y^{\rm GT})$ and $H(\cdot)$ are the mutual information and entropy, respectively; see \cite{strehl2002cluster} for more details.
  \item Similarity of estimated top $t$ essential features (Dice$_t$)\\
  We define similarity of estimated essential features as Dice index, defined as
\begin{equation*}
  {\rm Dice}_t(\mathcal{F}_t^\ast, \mathcal{F}_t^{\rm Pred}) = \frac{| \mathcal{F}_t^\ast \cap \mathcal{F}_t^{\rm Pred} |}{t},
\end{equation*}
where $\mathcal{F}_t^\ast$ and $\mathcal{F}_t^{\rm Pred}$ denote the estimated top $t$ essential features computed by centralized analysis that share the raw data and the intermediate DC analysis.
Note that $|\mathcal{F}_t^\ast| = |\mathcal{F}_t^{\rm Pred}| = t$.
\end{itemize}
\par
Numerical experiments were performed using Python and MATLAB \footnote{ Program codes are available from the corresponding author by reasonable request.}.
\subsection{Proof-of-concept for artificial dataset}
\begin{figure}[!t]
  \centering
  \subfloat[Training dataset]{
  \includegraphics[scale=0.35, bb = 50 250 545 602]{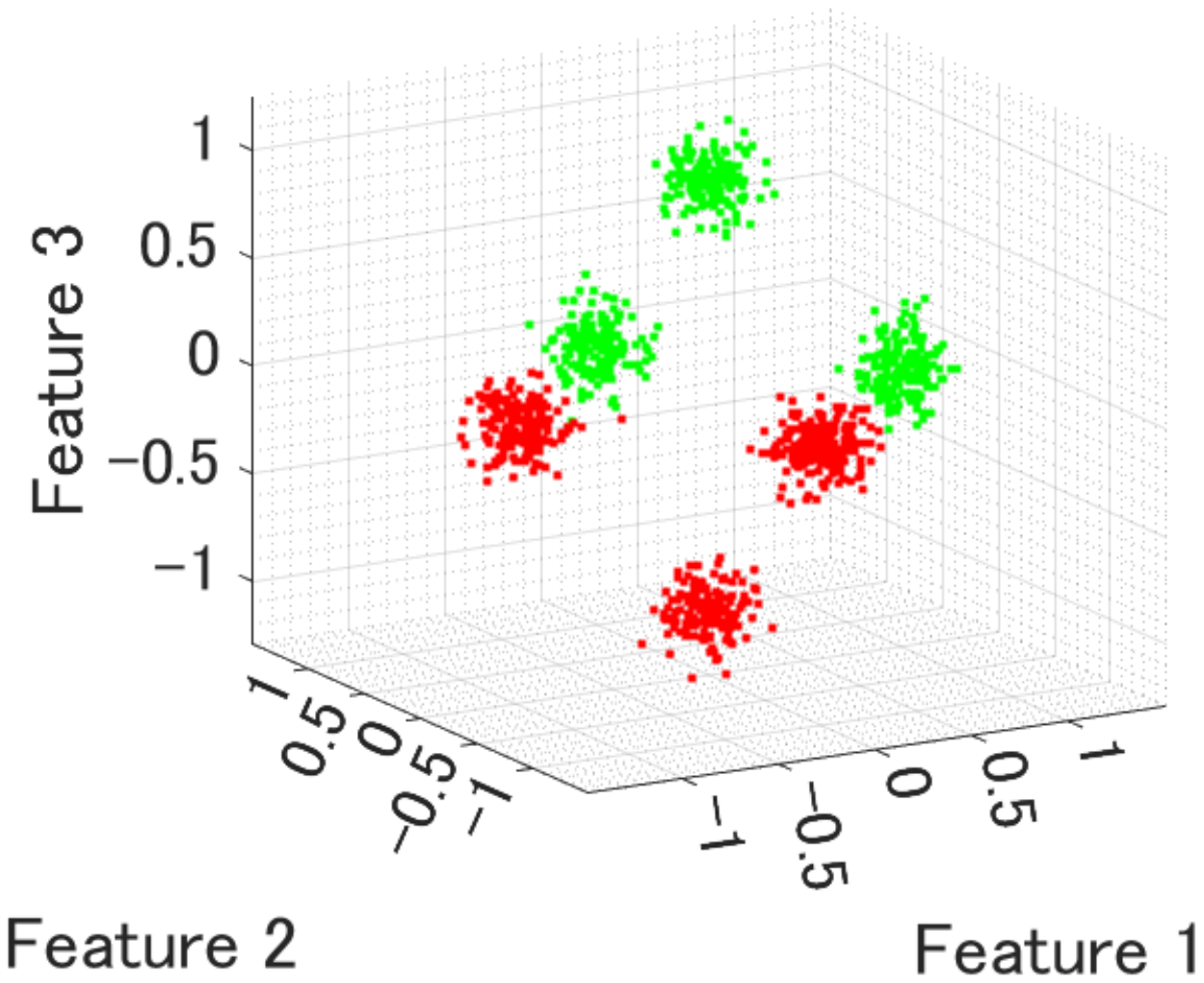}
  \includegraphics[scale=0.35, bb = 50 250 545 602]{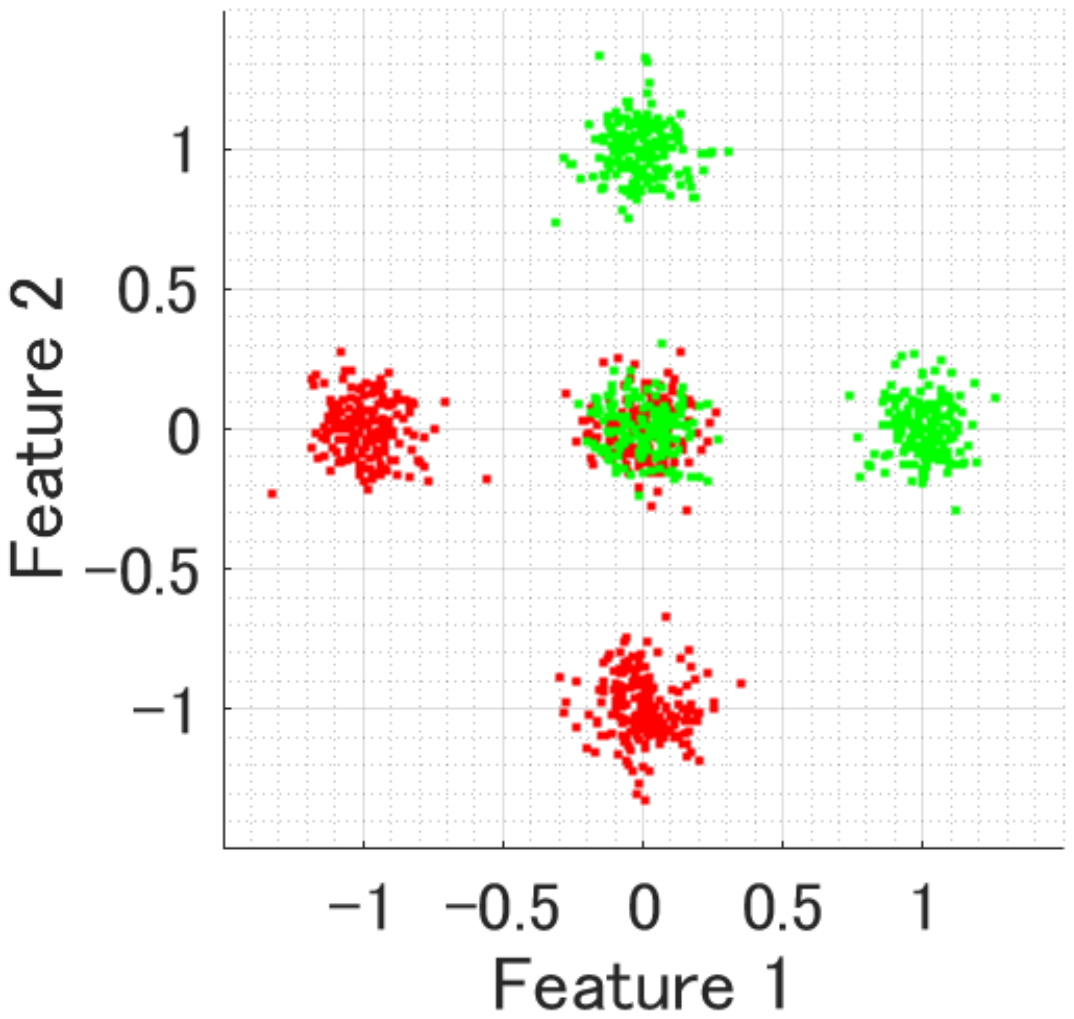}
}\\
  \subfloat[Test dataset]{
  \includegraphics[scale=0.35, bb = 50 250 545 602]{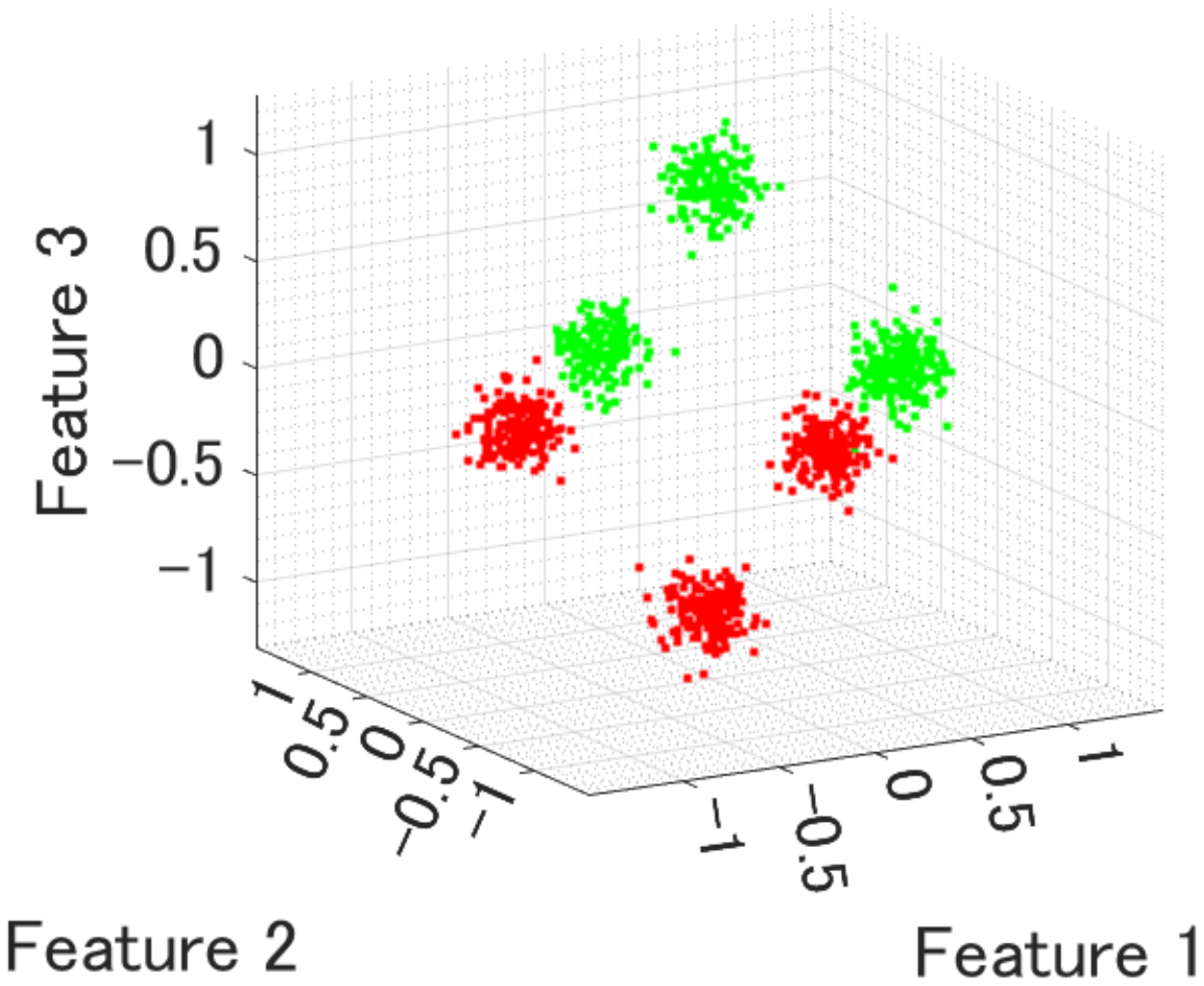}
  \includegraphics[scale=0.35, bb = 50 250 545 602]{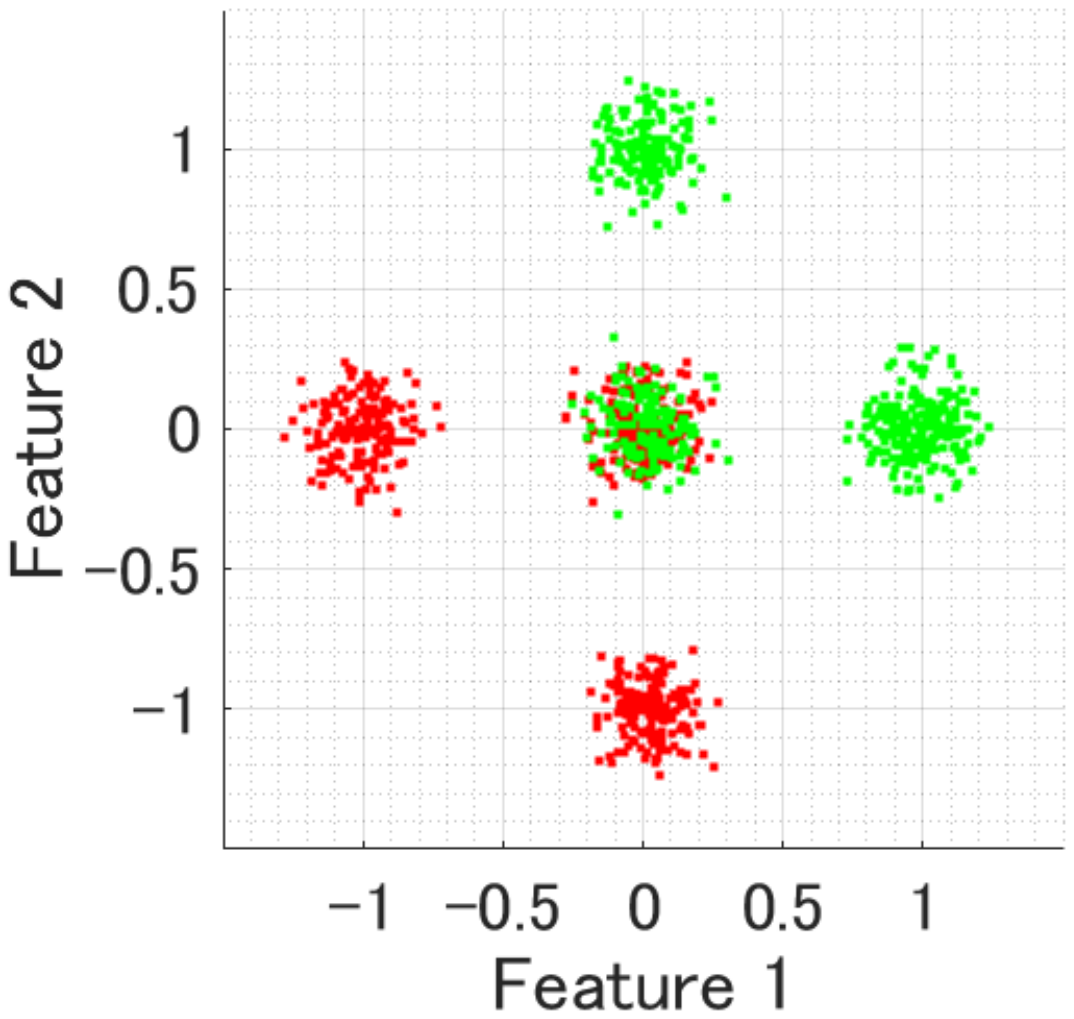}
}\\
  \subfloat[Public dataset]{
  \includegraphics[scale=0.35, bb = 50 250 545 602]{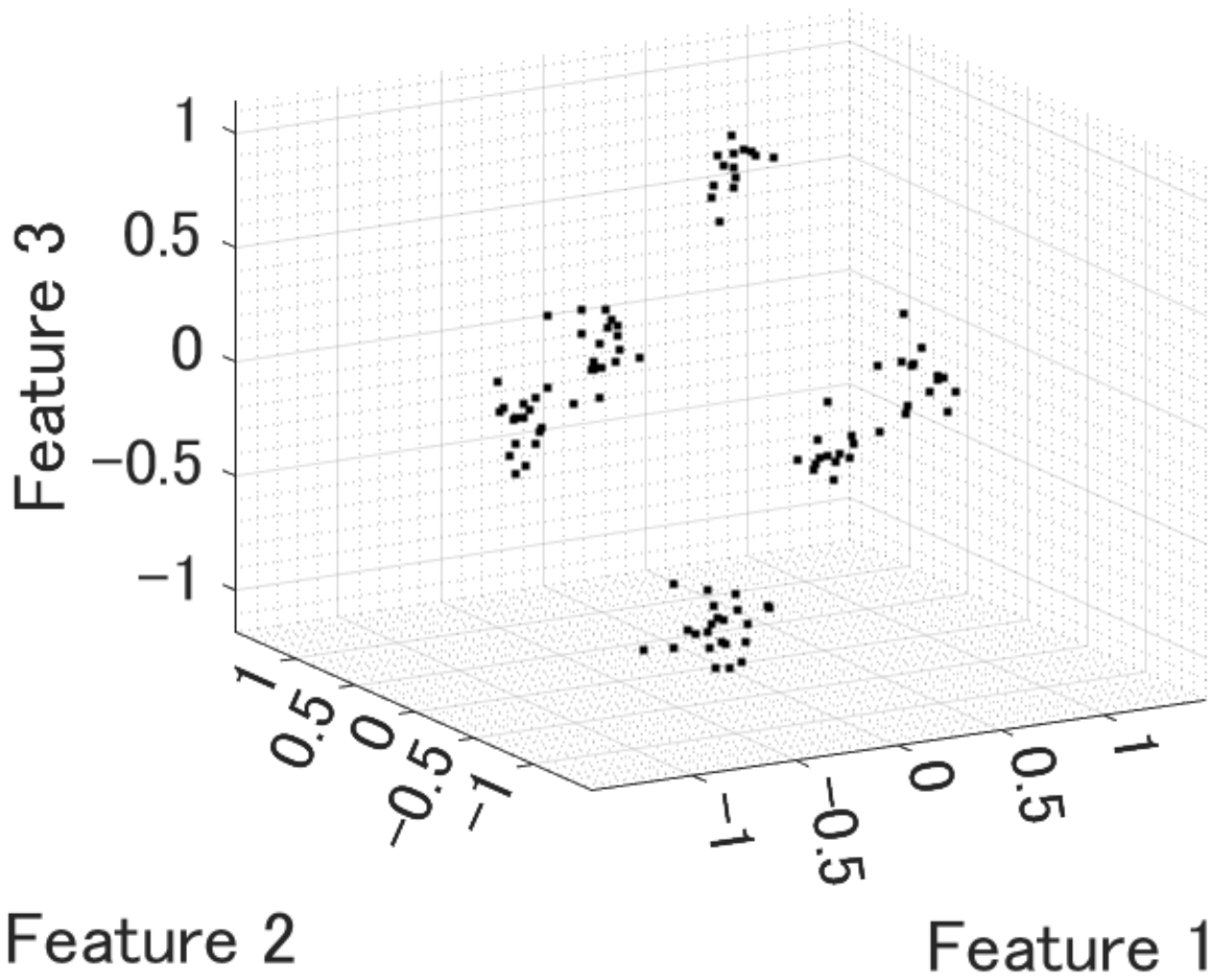}
  \includegraphics[scale=0.35, bb = 50 250 545 602]{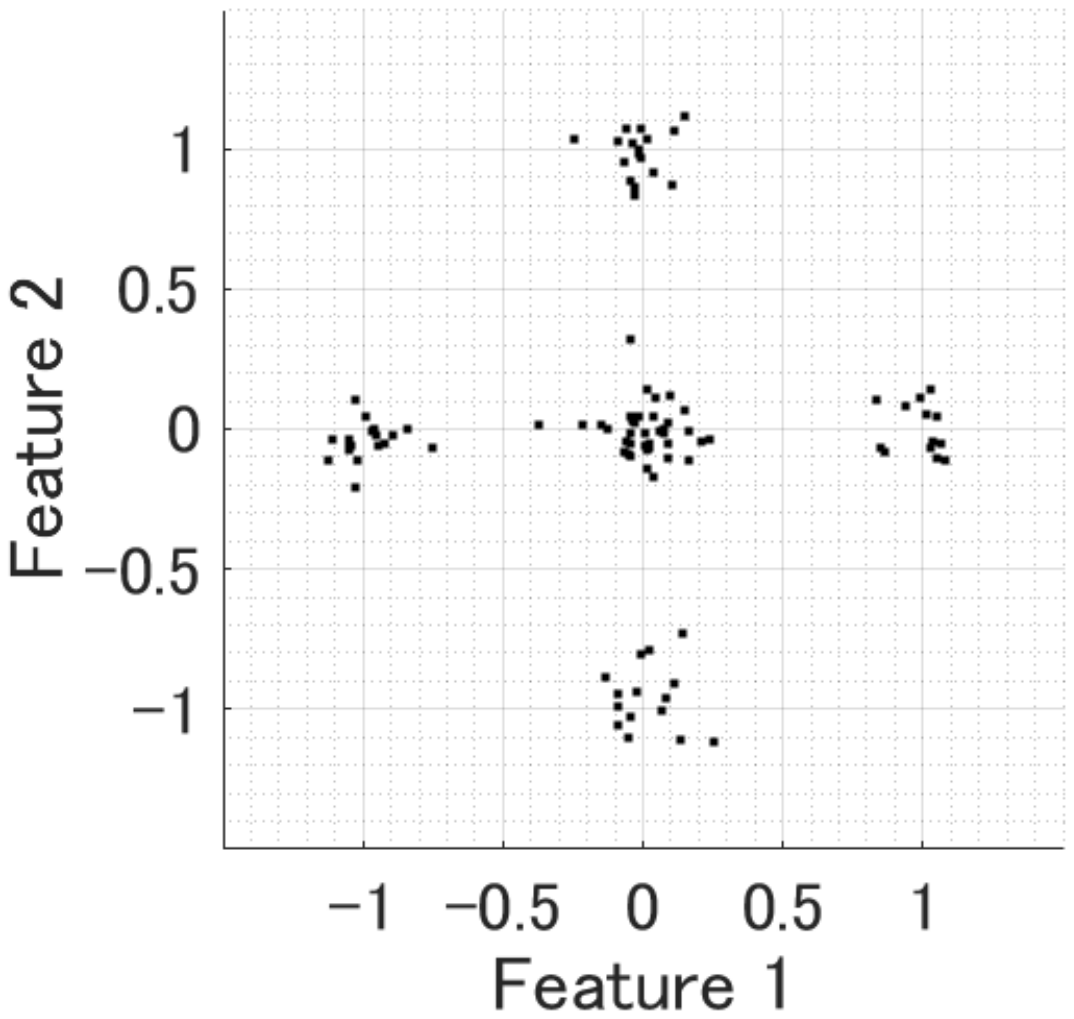}
}
\caption{Training, test, and public datasets for the artificial problem.}
\label{fig:train}
\end{figure}
\begin{figure}[!t]
  \centering
\subfloat[Random anchor data construction]{
  \includegraphics[scale=0.35, bb = 50 250 545 602]{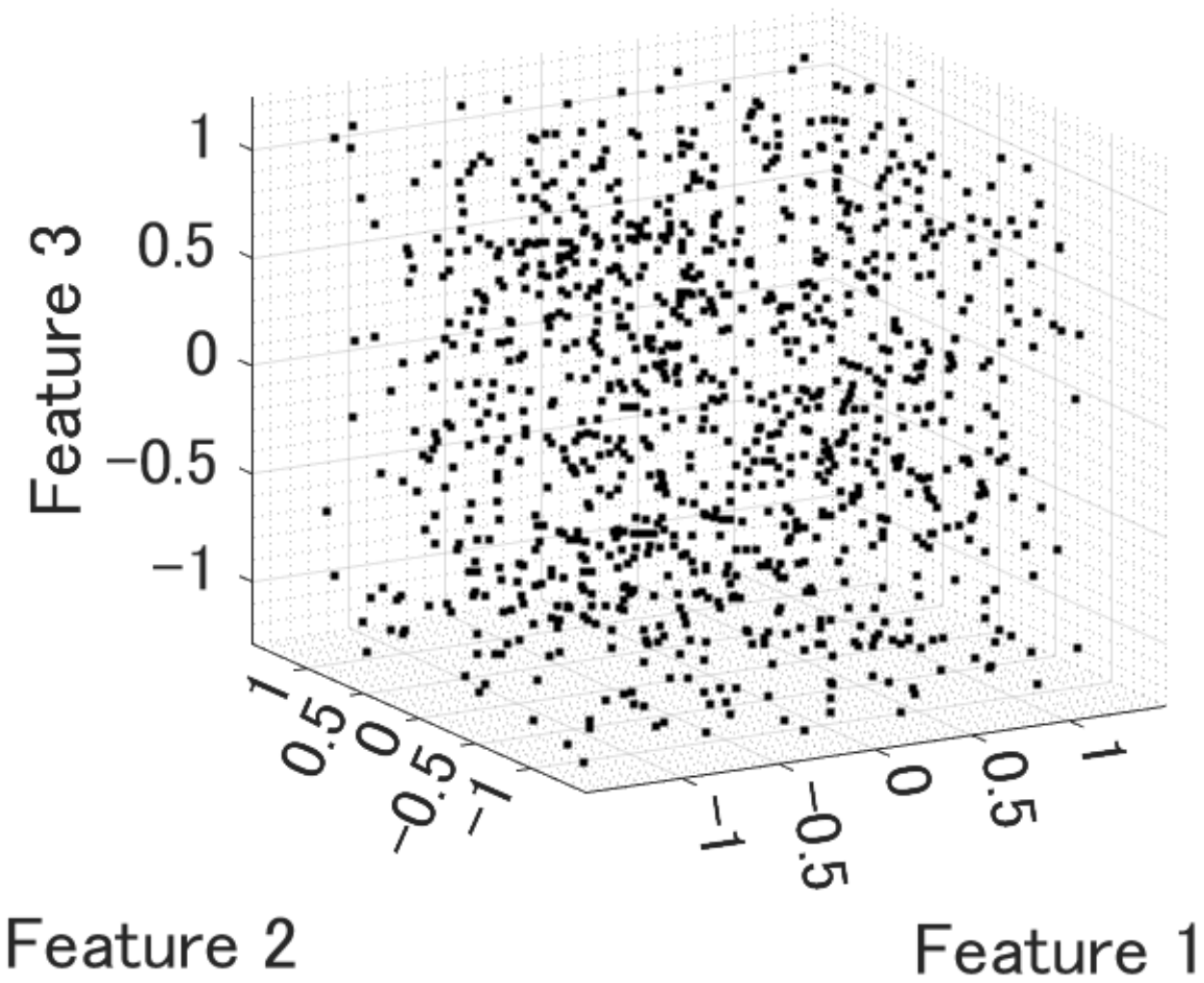}
  \includegraphics[scale=0.35, bb = 50 250 545 602]{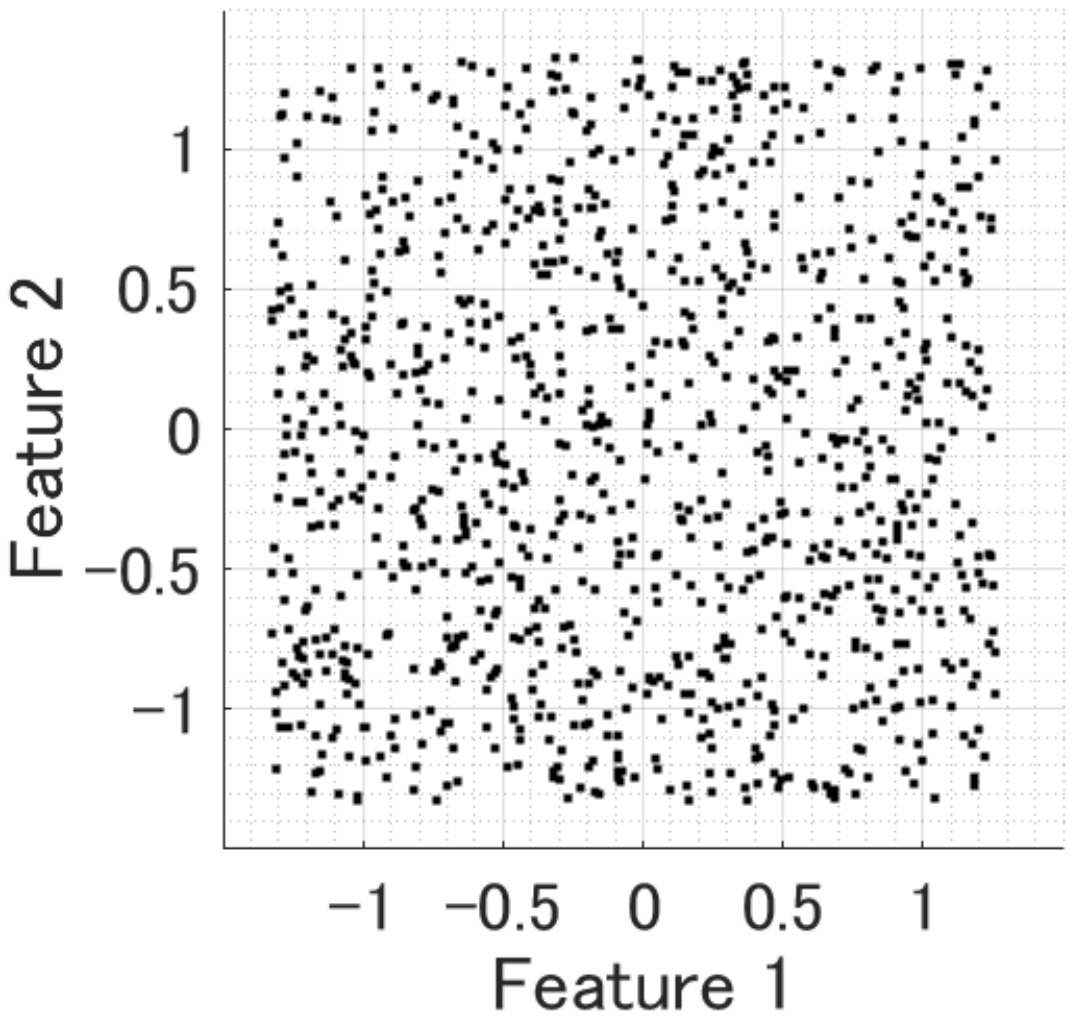}
} \\
\subfloat[TSVD-based anchor data construction]{
  \includegraphics[scale=0.35, bb = 50 250 545 602]{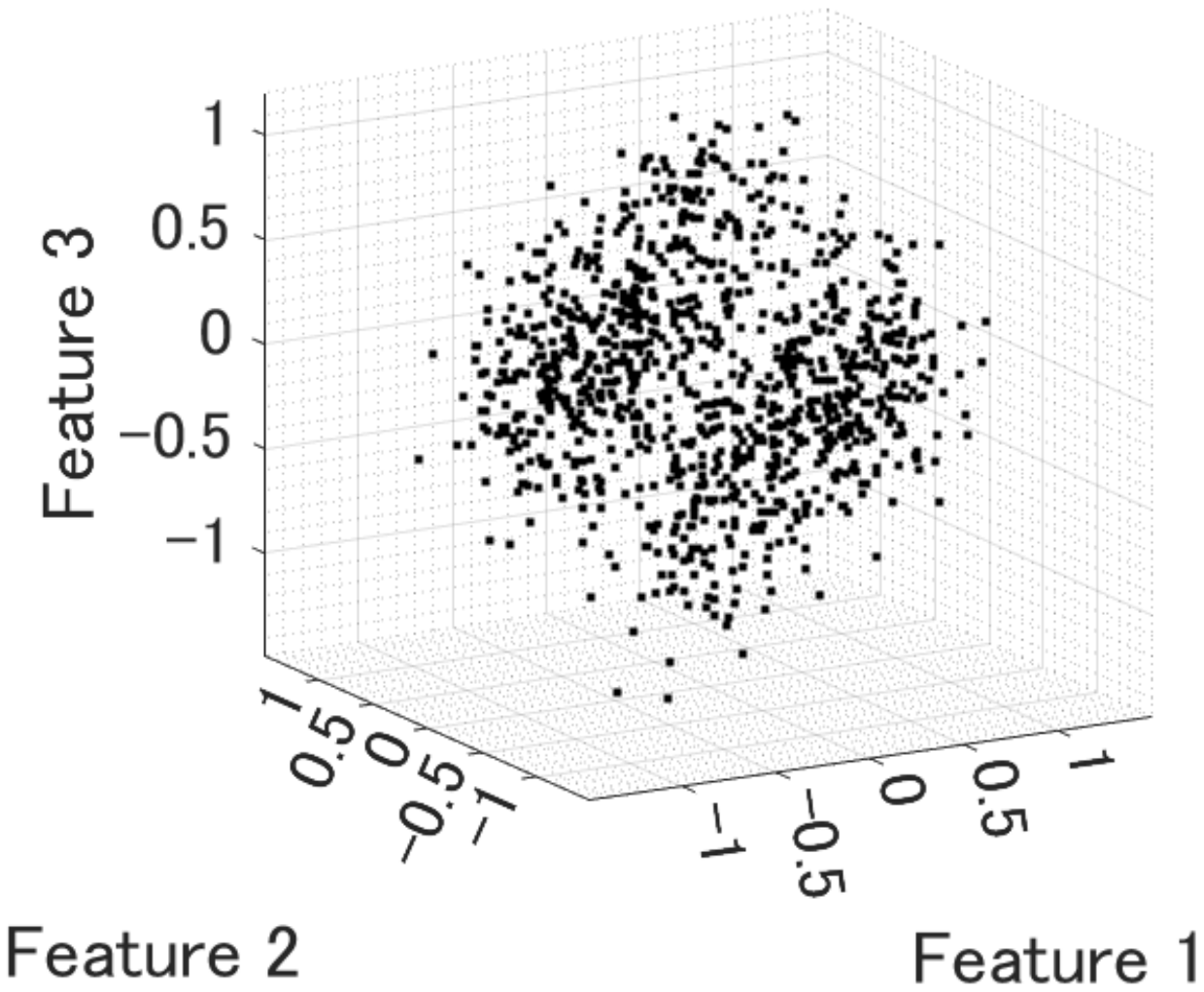}
  \includegraphics[scale=0.35, bb = 50 250 545 602]{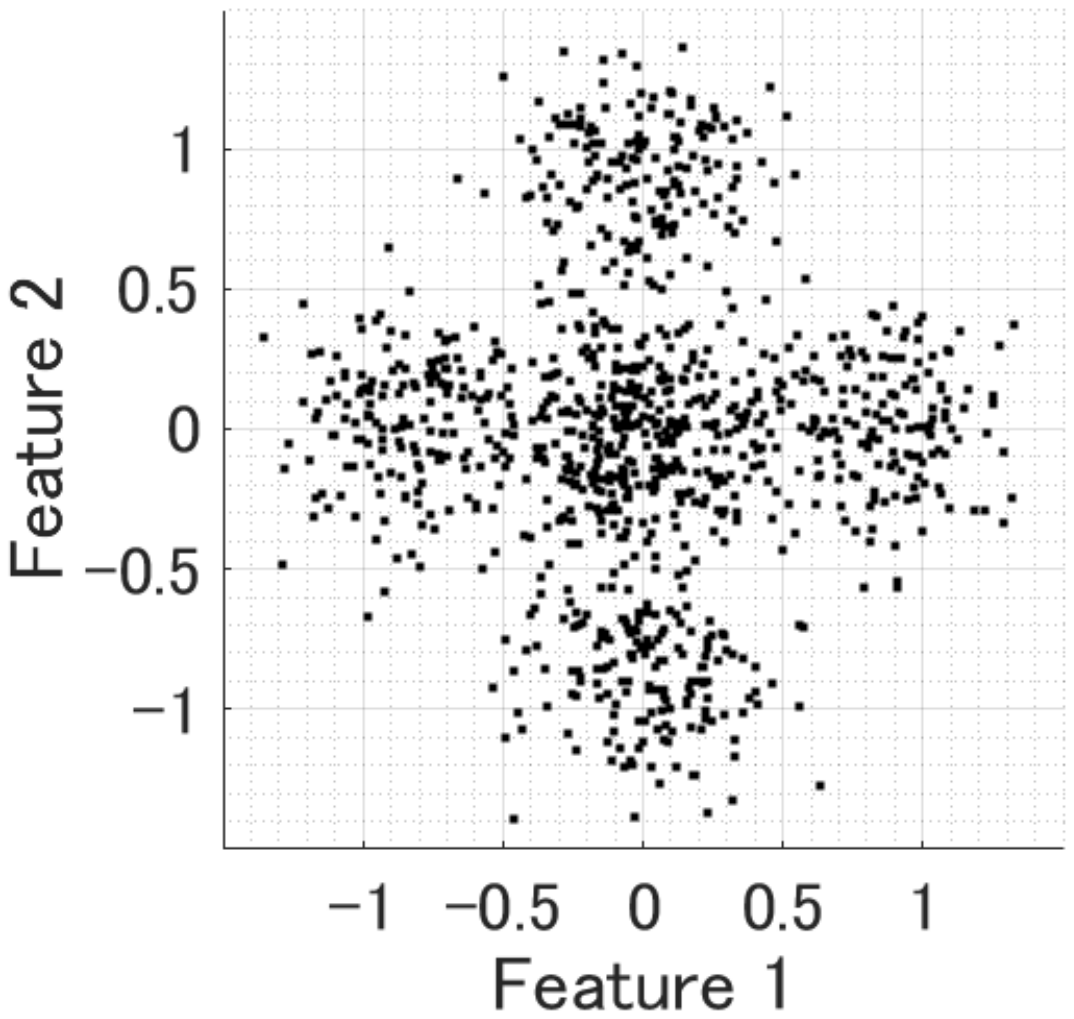}
} \\
\subfloat[SMOTE-based anchor data construction]{
  \includegraphics[scale=0.35, bb = 50 250 545 602]{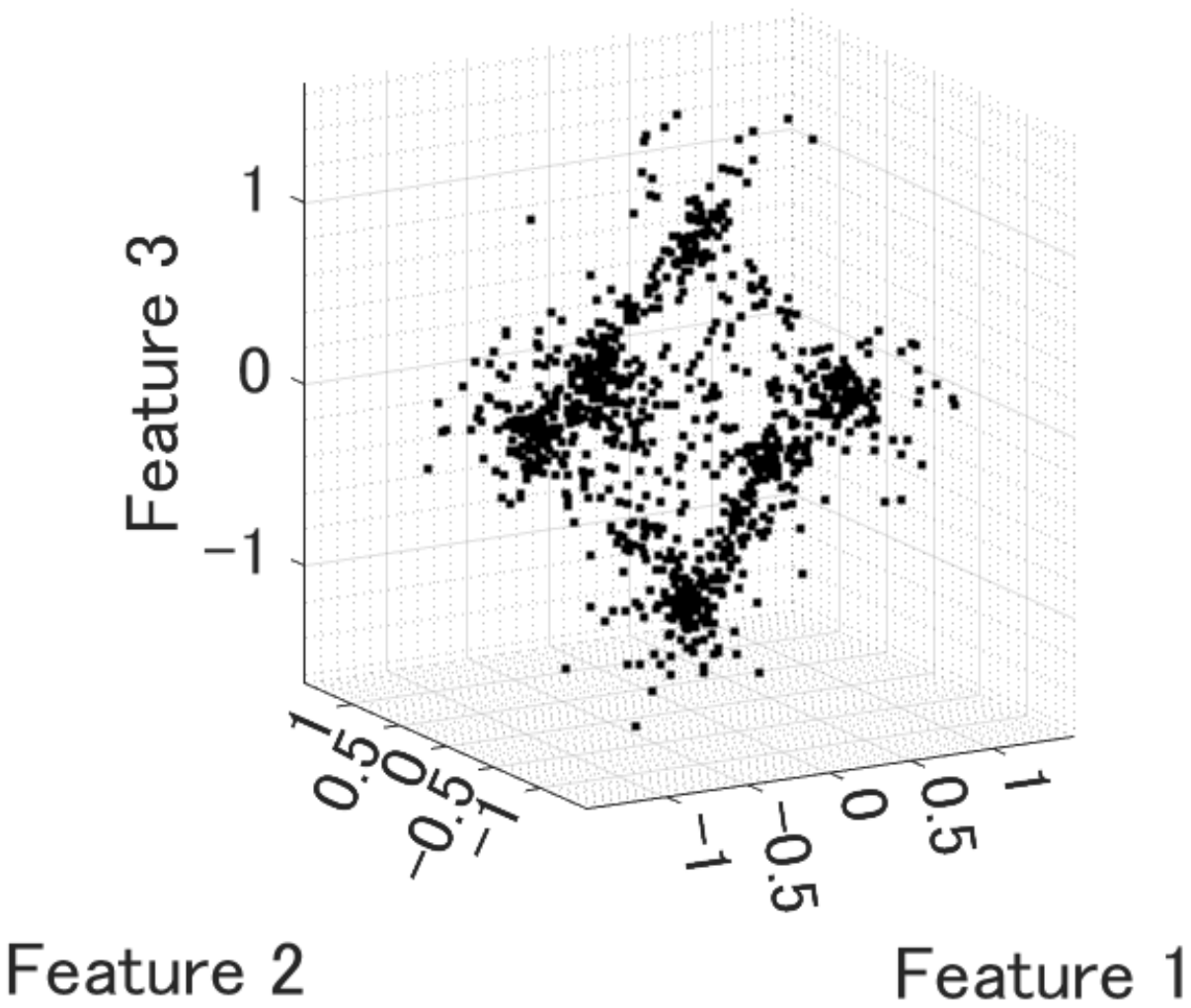}
  \includegraphics[scale=0.35, bb = 50 250 545 602]{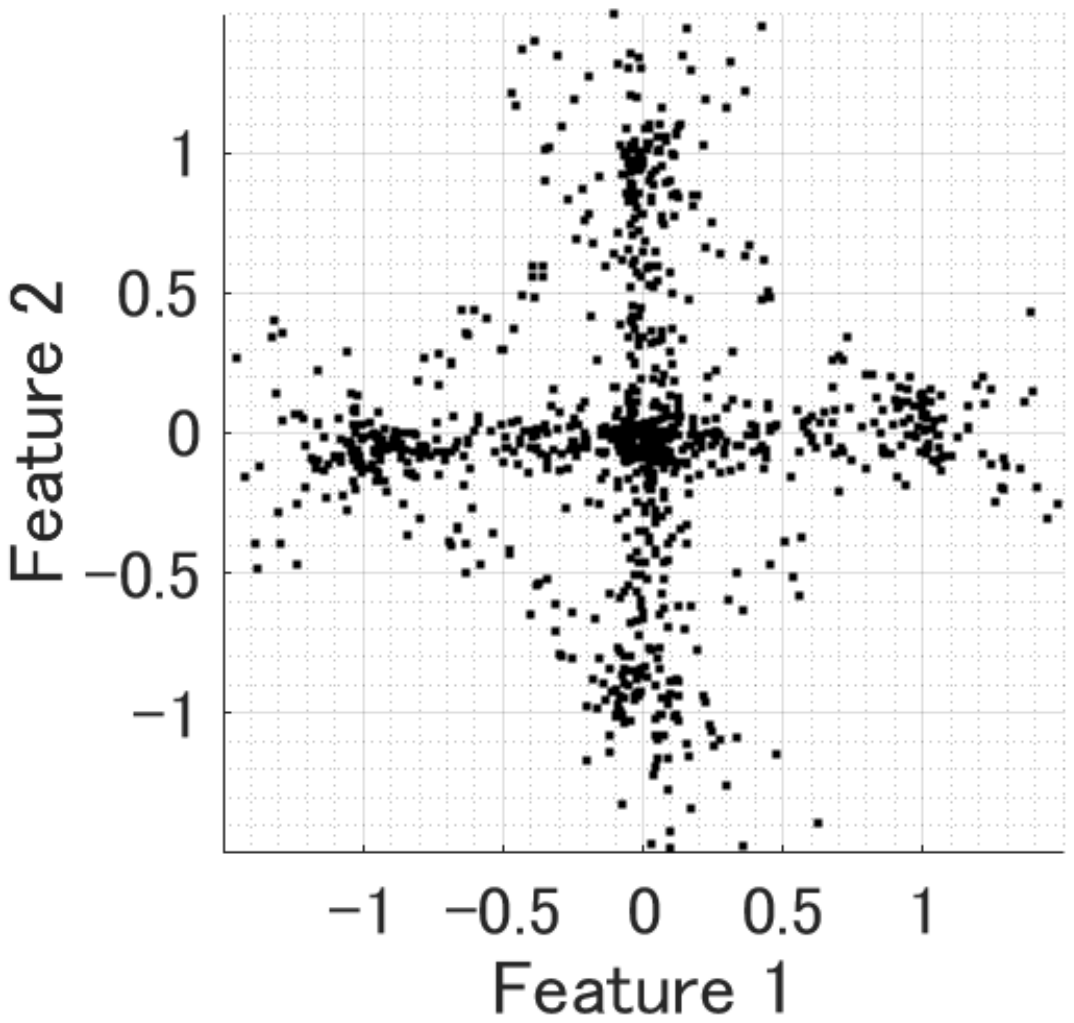}
} \\
\caption{Constructed anchor data for the artificial problem.}
\label{fig:anchor}
\end{figure}
As a proof of concept of the proposed method, we used a 20-dimensional artificial data for two-class classification.
Figure~\ref{fig:train} shows features 1, 2, and 3 of the training, test, and public datasets, where the numbers of samples are $(n_{\rm train}, n_{\rm test}, p) = (1000, 1000, 100)$.
The other 17 dimensions have random values.
Note that features 1, 2, and 3 are essential features for classification.
\par
We considered the case where the training dataset in Figure~\ref{fig:train}(a) is distributed into four parties, $c=d=2$, as
\begin{equation*}
  X = \left[ \begin{array}{cc}
      X_{1,1} & X_{1,2} \\
      X_{2,1} & X_{2,2} 
    \end{array}
  \right] \in \mathbb{R}^{1000 \times 20}, \quad
  X_{1,1}, X_{1,2}, X_{2,1}, X_{2,2} \in \mathbb{R}^{500 \times 10}.
\end{equation*}
For horizontal (sample) partitioning, we randomly partitioned the dataset into two groups.
For vertical (feature) partitioning, $X_{1,1}, X_{2,1}$ have odd features, and $X_{1,2}, X_{2,2}$ have even features.
Note that the essential features (1, 2, and 3) are partitioned into two groups.
\par
For the DC analysis, we set the dimensionality of intermediate representations to $\widetilde{m}_{i,j} = 5$ for all parties and the number of anchor data to $r=1000$.
We used the ridge regression for analyzing the collaboration representation (Step 9 in Algorithm 1) and the decision tree as interpretable model (Step 13 in Algorithm~\ref{alg:IDC}).
We set five as the maximum number of branch node splits. 
For the TSVD-based method, we set the rank of TSVD to $3$.
For the proposed SMOTE-based method, we set $(k,\alpha) = (25,1.5)$.
\par
As numerical results, we show constructed the anchor data using each method in Figure~\ref{fig:anchor}.
We also show the average and standard error of recognition performance (NMI, ACC, and Dice$_3$) and data confidentiality (AMD(raw) and AMD(anc)) for each method across 20 trials in Table~\ref{table:ex1_result}.
\par
We note that unlike the random anchor data construction, the TSVD-based and SMOTE-based methods generated anchor data along the distribution of the raw data.
Regarding recognition performance (NMI, ACC, and Dice$_3$ in Table~\ref{table:ex1_result}), {\bf DC(SMOTE)} correctly finds the top $3$ essential features and shows high recognition performance comparable to {\bf DC(raw)} and {\bf Centralized}.
Specifically, {\bf DC(SMOTE)} achieved 7 percentage point (NMI), 4 percentage point (ACC), and 7 percentage point (Dice$_3$) performance improvements over {\bf DC(TSVD)}, respectively.
Note that {\bf DC(TSVD)} performs better performance than {\bf Local} and {\bf DC(rand)}, but worse than {\bf DC(SMOTE)}.
Regarding data confidentiality (AMD(raw) and AMD(anc) in Table~\ref{table:ex1_result}), {\bf DC(SMOTE)} shows larger values than {\bf DC(TSVD)}, which indicates that {\bf DC(SMOTE)} has better data confidentiality than {\bf DC(TSVD)}.
\par
Overall, {\bf DC(SMOTE)} demonstrates a high level of both recognition accuracy and data confidentiality for the artificial dataset.
\begin{table}[!t]
  \small
    \centering
    \caption{Prediction results of each method for the artificial dataset.}
    \label{table:ex1_result}
    \begin{tabular}{lccccc}
      \toprule
      \multicolumn{1}{c}{Method} & \multicolumn{3}{c}{Recognition performance} & \multicolumn{2}{c}{Data confidentiality} \\ \cmidrule(rl){2-4} \cmidrule(rl){5-6}
      & NMI & ACC & Dice$_3$ & AMD(raw) & AMD(anc) \\ \midrule
      Centralized  & 0.99$\pm$0.00 & 1.00$\pm$0.00 & 1.00$\pm$0.00 & & \\
      Local        & 0.26$\pm$0.02 & 0.75$\pm$0.01 & 0.50$\pm$0.02 &	& \\ \midrule
      DC(raw)     & 0.97$\pm$0.01 & 1.00$\pm$0.00 & 1.00$\pm$0.00 &	0.00$\pm$0.00 & 0.00$\pm$0.00 \\ \midrule
      DC(rand)    & 0.52$\pm$0.05 & 0.85$\pm$0.02 & 0.92$\pm$0.03 &	2.15$\pm$0.00 & 2.23$\pm$0.00 \\
      DC(TSVD)    & 0.84$\pm$0.07 & 0.95$\pm$0.02 & 0.93$\pm$0.03 &	1.87$\pm$0.00 & 1.58$\pm$0.00 \\
      DC(SMOTE)   & 0.97$\pm$0.01 & 0.99$\pm$0.00 & 1.00$\pm$0.00 &	2.02$\pm$0.00 & 1.90$\pm$0.01 \\
        \bottomrule
    \end{tabular}
\end{table}
\subsection{Evaluation regarding recognition performance and data confidentiality on a credit rating dataset}
Here, we evaluate the trade-off between recognition performance and data confidentiality of each method for a credit rating dataset ``CreditRating\_Historical.dat'' from the MATLAB Statistics and Machine Learning Toolbox.
The dataset contains five financial ratios, i.e., Working capital / Total Assets (WC\_TA), Retained Earnings / Total Assets (RE\_TA), Earnings Before Interests and Taxes / Total Assets (EBIT\_TA), Market Value of Equity / Book Value of Total Debt (MVE\_BVTD), and Sales / Total Assets (S\_TA), and industry sector labels from 1 to 12 for 3932 customers.
The dataset also includes credit ratings from ``AAA'' to ``CCC'' for all customers.
The categorical variable ``Industry sector labels'' was transformed into 12 dimensional dummy variables.
Note that this dataset is simulated, not real.
\par
We sought to predict credit rating using the five financial ratios and industry sector labels.
We considered the case where the training dataset with 3,000 samples is distributed into four parties, $c=d=2$, as
\begin{equation*}
  X = \left[ \begin{array}{cc}
      X_{1,1} & X_{1,2} \\
      X_{2,1} & X_{2,2} 
    \end{array}
  \right] \in \mathbb{R}^{3000 \times 17}
\end{equation*}
with
\begin{equation*}
  X_{1,1}, X_{2,1} \in \mathbb{R}^{1500 \times 3}, \quad
  X_{1,2}, X_{2,2} \in \mathbb{R}^{1500 \times 14},
\end{equation*}
where, $X_{1,1}, X_{2,1}$ have the 1st group of features ``WC\_TA'', ``RE\_TA'', and ``EBIT\_TA'', and $X_{1,2}, X_{2,2}$ have the 2nd group of features ``MVE\_BVTD'', ``S\_TA'', and ``Industry sector labels'' as features.
\par
For the DC analysis, we set the dimensionality of intermediate representations to $\widetilde{m}_{i,j} = m_j - 1$ for all parties and the number of anchor data to $r=2500$.
We used the XGBoost method with default parameters in Python for analyzing the collaboration representation (Step 9 in Algorithm~\ref{alg:IDC}) and for interpretable model (Step 13 in Algorithm~\ref{alg:IDC}).
For the TSVD-based method, we set the rank of TSVD to 1--3.
For the proposed SMOTE-based method, we set $(k,\alpha) = (99,1.5)$.
\par
As numerical results, we show the average and standard error of recognition performance (NMI, ACC, and Dice$_2$) and data confidentiality (EMD, AMD(raw), and AMD(anc)) for each method across ten trials in Table~\ref{table:ex2_result1} and \ref{table:ex2_result2}, respectively.
\par
Table~\ref{table:ex2_result1} shows that the DC methods correctly finds the top $2$ essential features except {\bf DC(TSVD)} with rank $1$.
Additionally, the proposed {\bf DC(SMOTE)} shows a high recognition performance comparable to {\bf DC(raw)} and {\bf Centralized}.
However, regarding data confidentiality (Table~\ref{table:ex2_result2}), {\bf DC(SMOTE)} shows larger AMD(raw), smaller AMD(anc), and almost the same EMD compared with {\bf DC(TSVD)} with ranks 2 and 3.
This result indicates that {\bf DC(SMOTE)} shows almost the same data confidentiality as {\bf DC(TSVD)} with rank 2 and 3.
\begin{table}[!t]
  \small
    \centering
    \caption{Recognition performance of each method for the CreditRating\_Historical.dat.}
    \label{table:ex2_result1}
    \begin{tabular}{lccc}
      \toprule
      \multicolumn{1}{c}{Method} & NMI & ACC & Dice$_2$ \\ \midrule
Centralized  & 0.61$\pm$0.01&0.74$\pm$0.01&1.00	\\
Local        & 0.46$\pm$0.01&0.60$\pm$0.00&			\\
DC(raw)     & 0.61$\pm$0.01&0.74$\pm$0.01&1.00	\\
DC(rand)    & 0.51$\pm$0.04&0.54$\pm$0.07&1.00	\\
DC(TSVD) rank$=$1 & 0.45$\pm$0.07&0.52$\pm$0.08&0.55	\\
DC(TSVD) rank$=$2 & 0.58$\pm$0.02&0.71$\pm$0.02&1.00	\\
DC(TSVD) rank$=$3 & 0.60$\pm$0.01&0.73$\pm$0.02&1.00	\\
DC(SMOTE)   & 0.61$\pm$0.01&0.74$\pm$0.01&1.00	\\
        \bottomrule
    \end{tabular}
\end{table}
\begin{table}[!t]
  \small
    \centering
    \caption{Data confidentiality of each method for the CreditRating\_Historical.dat.}
    \label{table:ex2_result2}
    \begin{tabular}{lccc}
      \toprule
      \multicolumn{1}{c}{Method} & EMD & AMD(raw) & AMD(anc) \\ \midrule
DC(raw)    &  0.00$\pm$0.00&0.00$\pm$0.00&0.00$\pm$0.00 \\
DC(rand)   & 52.84$\pm$9.49&1.53$\pm$0.17&8.40$\pm$2.49 \\
DC(TSVD) rank$=$1 & 2.12$\pm$0.07&1.72$\pm$0.13&0.63$\pm$0.01 \\
DC(TSVD) rank$=$2 & 1.17$\pm$0.04&0.28$\pm$0.00&0.85$\pm$0.02 \\
DC(TSVD) rank$=$3 & 1.17$\pm$0.04&0.26$\pm$0.00&0.85$\pm$0.02 \\
DC(SMOTE)  & 1.15$\pm$0.26&0.36$\pm$0.07&0.50$\pm$0.17 \\
        \bottomrule
    \end{tabular}
\end{table}
\subsection{Evaluation regarding recognition performance for income dataset with two data distribution scenarios}
Here, we evaluate the recognition performance of each method for an income dataset ``Adult'' from the UCI Machine Learning Repository.
The prediction task is to determine whether a person makes more than \$50,000 per year.
We used five continuous features and seven categorical features, excluding ``fnlwgt``and ``ducation''.
The seven categorical variables were transformed into 86-dimensional dummy variables.
\par
We considered the case where the training dataset with 30,000 samples is distributed into four parties, $c=d=2$, as
\begin{equation*}
  X = \left[ \begin{array}{cc}
      X_{1,1} & X_{1,2} \\
      X_{2,1} & X_{2,2} 
    \end{array}
  \right] \in \mathbb{R}^{15,000 \times 91}
\end{equation*}
with
\begin{equation*}
  X_{1,1}, X_{2,1} \in \mathbb{R}^{15,000 \times m_1}, \quad
  X_{1,2}, X_{2,2} \in \mathbb{R}^{15,000 \times m_2},
\end{equation*}
where samples were randomly partitioned, and features were partitioned based on two distribution scenarios:
\begin{itemize}
  \item {\sf Artificial:} \\ 91 features are partitioned into two groups using even or odd indices.
    Here, $m_1 = 46$ and $m_2 = 45$.

  \item {\sf Feature type:} \\ 91 features are partitioned into two groups using continuous or categorical features.
    Here, $m_1 = 5$ and $m_2 = 86$.
\end{itemize}
\par
For the DC analysis, we set the dimensionality of intermediate representations to $\widetilde{m}_{i,j} = m_j - 1$ for all parties and the number of anchor data to $r=2500$.
We used the XGBoost method with default parameters in Python for analyzing the collaboration representation (Step 9 in Algorithm~\ref{alg:IDC}) and for interpretable model (Step 13 in Algorithm~\ref{alg:IDC}).
For the TSVD-based method, we set the rank of TSVD to 2 and $m_j-1$.
For the proposed SMOTE-based method, we set $(k,\alpha) = (99,1.5)$.
\begin{table}[!t]
  \small
    \centering
    \caption{Recognition performance of each method for Adult with {\sf Artificial} data partitioning scenario.}
    \label{table:ex3_result1}
    \begin{tabular}{lccc}
      \toprule
      \multicolumn{1}{c}{Method} & NMI & ACC & Dice$_5$ \\ \midrule
      Centralized           & 0.34$\pm$0.00 & 0.87$\pm$0.00 & 1.00$\pm$0.00 \\
      Local                 & 0.22$\pm$0.00 & 0.83$\pm$0.00 & 0.50$\pm$0.00 \\
      DC(raw)               & 0.33$\pm$0.00 & 0.87$\pm$0.00 & 1.00$\pm$0.00 \\
      DC(rand)              & 0.12$\pm$0.05 & 0.72$\pm$0.12& 0.24$\pm$0.12 \\
      DC(TSVD) rank$=$2     & 0.00$\pm$0.01 & 0.76$\pm$0.00& 0.46$\pm$0.18 \\
      DC(TSVD) rank$=m_j-1$ & 0.00$\pm$0.01 & 0.76$\pm$0.00 & 0.54$\pm$0.13 \\
      DC(SMOTE)             & 0.27$\pm$0.02 & 0.85$\pm$0.01 & 0.92$\pm$0.10 \\
        \bottomrule
    \end{tabular}
\end{table}
\begin{table}[!t]
  \small
    \centering
    \caption{Recognition performance of each method for Adult with {\sf Feature type} data partitioning scenario.}
    \label{table:ex3_result2}
    \begin{tabular}{lccc}
      \toprule
      \multicolumn{1}{c}{Method} & NMI & ACC & Dice$_5$ \\ \midrule
Centralized           & 0.34$\pm$0.00 & 0.87$\pm$0.00& 1.00$\pm$0.00 \\
Local                 & 0.22$\pm$0.00 & 0.83$\pm$0.00& 0.50$\pm$0.00 \\
DC(raw)               & 0.32$\pm$0.00 & 0.87$\pm$0.00& 0.98$\pm$0.06 \\
DC(rand)              & 0.13$\pm$0.04 & 0.78$\pm$0.03& 0.40$\pm$0.13 \\
DC(TSVD) rank$=$2     & 0.22$\pm$0.01 & 0.82$\pm$0.00& 0.46$\pm$0.09 \\
DC(TSVD) rank$=m_j-1$ & 0.21$\pm$0.01 & 0.81$\pm$0.01& 0.48$\pm$0.10 \\
DC(SMOTE)             & 0.26$\pm$0.03 & 0.85$\pm$0.01& 0.80$\pm$0.13 \\
        \bottomrule     
    \end{tabular}
\end{table}
\par
Tables~\ref{table:ex3_result1} and \ref{table:ex3_result2} show that the proposed {\bf DC(SMOTE)} has a high recognition performance comparable to {\bf DC(raw)} and {\bf Centralized} compared with {\bf DC(rand)} and {\bf DC(TSVD)}.
Additionally, the proposed {\bf DC(SMOTE)} finds the top $5$ essential features with higher rates than {\bf DC(rand)} and {\bf DC(TSVD)} even for {\sf Feature type} data distribution scenario. 
Specifically, {\bf DC(SMOTE)} achieved 9 percentage point (ACC) and 38 percentage point (Dice$_5$) performance improvements over {\bf DC(TSVD)} for {\sf Artificial} data distribution scenario, and 4 percentage point (ACC) and 32 percentage point (Dice$_5$) performance improvements over {\bf DC(TSVD)} for {\sf Feature type} data distribution scenario.
\subsection{Evaluation for parameter dependency}
To evaluate the parameter dependency of the proposed method, we used the ``CreditRating\_Historical.dat'' and ``Adult''.
We evaluated the recognition performance of the proposed method by varying $k = 1, 5, 10, \dots, 95, 99$ and $\alpha = 0.1, 0.2, \dots, 5.0$.
Other parameters were set as in Sections~4.2 and 4.3.
\par
\begin{figure}[!t]
  \centering
  \subfloat[``CreditRating\_Historical.dat'']{
  \includegraphics[scale=0.35, bb =  50 250 545 602]{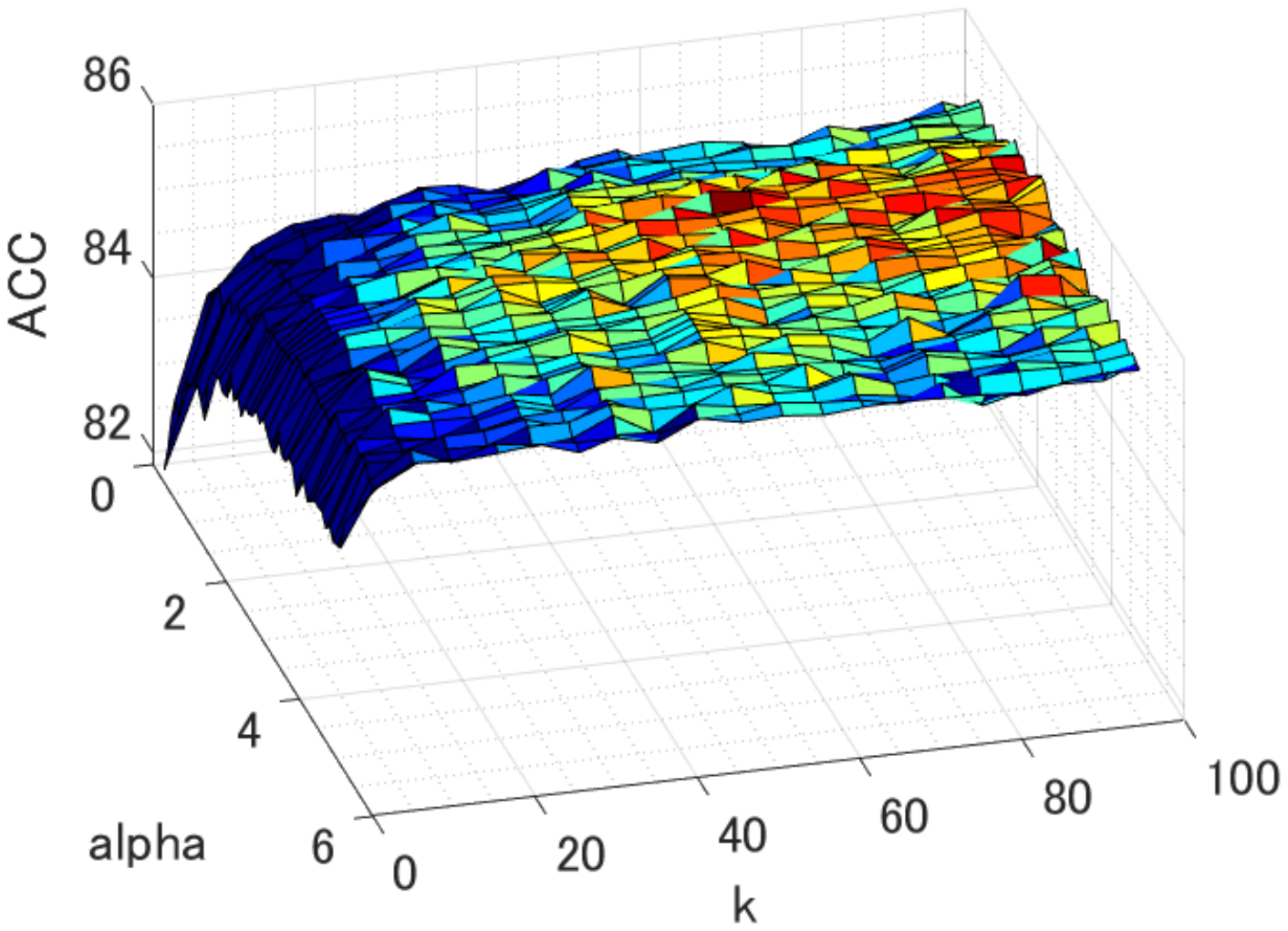}
}
  \subfloat[``Adult'']{
  \includegraphics[scale=0.35, bb =  50 250 545 602]{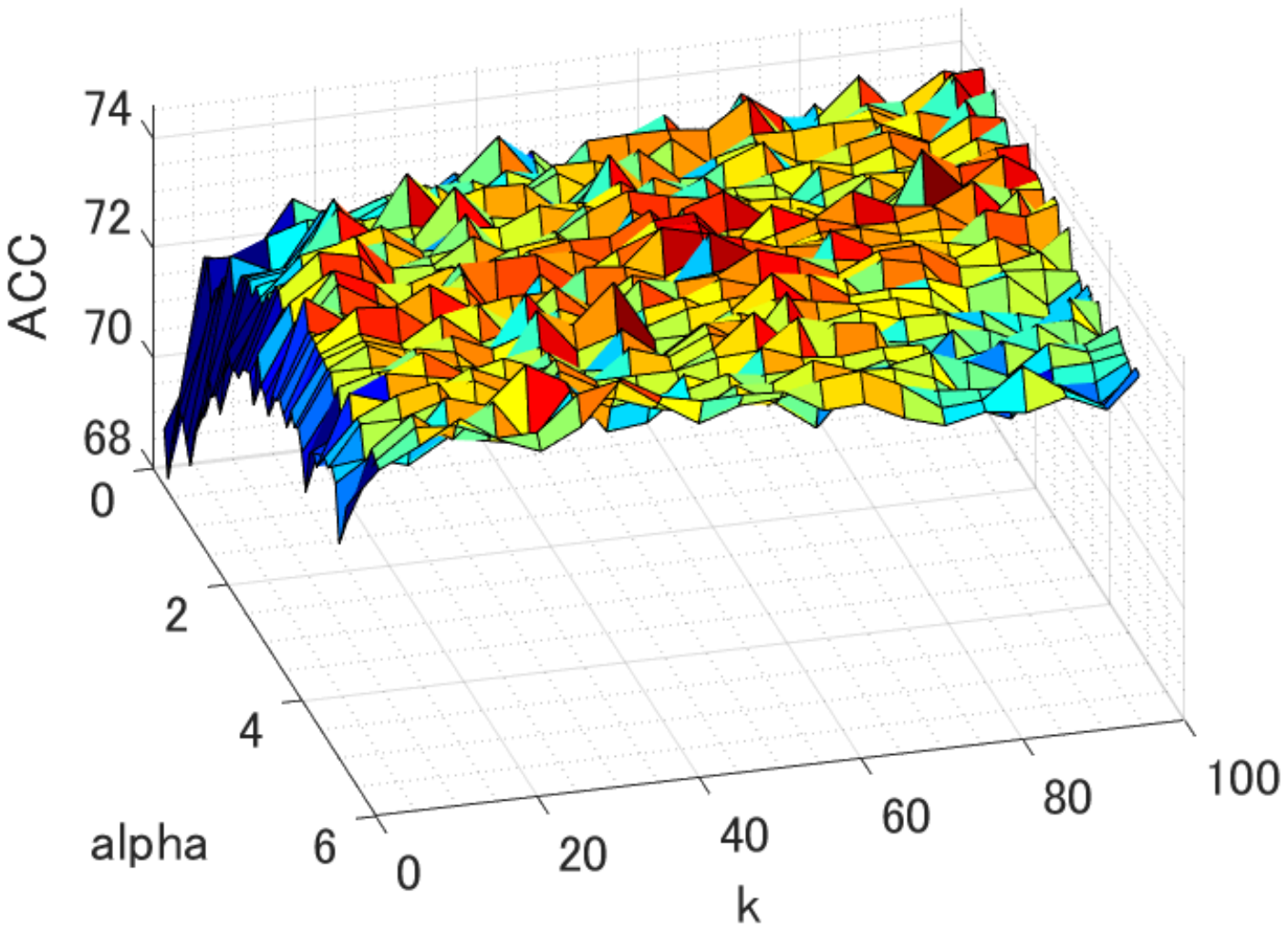}
}
\caption{Parameter dependency of the proposed method.}
\label{fig:ex4}
\end{figure}

Figure~\ref{fig:ex4} shows the average ACC of the proposed method.
This result implies that the proposed method with the standard parameter settings of the conventional SMOTE ($\alpha = 1$ and a small $k$) provides low accuracy.
Instead, the accuracy could be improved by the extension using a large $\alpha$ and a large $k$.
\subsection{Performance evaluation on real-world data}
\begin{table}[!t]
  \footnotesize
  \caption{Recognition performance (average $\pm$ standard error) for real-world problems (the first five of ten datasets).}
\label{table:result}
\begin{center}
\begin{tabular}{lclcccc} 
\toprule
\multicolumn{1}{c}{Dataset} &  & \multicolumn{1}{c}{Method} &  & NMI &  & ACC  \\ \cmidrule{1-1} \cmidrule{1-1} \cmidrule{3-3} \cmidrule{5-5} \cmidrule{7-7}
Carcinom     &  & Centralized &  & $0.66\pm0.03$ & & $54.58\pm3.06$  \\
$\;\;m=9182$ &  & Local       &  & $0.50\pm0.04$ & & $40.67\pm3.12$  \\
$\;\;n\;=174$&  & DC(TSVD)    &  & $0.66\pm0.04$ & & $54.06\pm5.30$  \\
             &  & DC(SMOTE)   &  & $0.66\pm0.03$ & & $56.16\pm3.53$  \\ \cmidrule{1-1} \cmidrule{3-3} \cmidrule{5-5} \cmidrule{7-7}
CLL-SUB-111  &  & Centralized &  & $0.22\pm0.03$ & & $60.40\pm2.86$  \\
$\;\;m=11340$&  & Local       &  & $0.16\pm0.02$ & & $56.81\pm1.64$  \\
$\;\;n\;=111$&  & DC(TSVD)    &  & $0.29\pm0.08$ & & $52.06\pm5.43$  \\
             &  & DC(SMOTE)   &  & $0.08\pm0.02$ & & $56.80\pm3.15$  \\ \cmidrule{1-1} \cmidrule{3-3} \cmidrule{5-5} \cmidrule{7-7}
GLA-BRA-180  &  & Centralized &  & $0.37\pm0.05$ & & $62.22\pm3.74$  \\
$\;\;m=49151$&  & Local       &  & $0.28\pm0.02$ & & $55.74\pm1.75$  \\
$\;\;n\;=180$&  & DC(TSVD)    &  & $0.33\pm0.04$ & & $61.67\pm2.41$  \\
             &  & DC(SMOTE)   &  & $0.29\pm0.06$ & & $59.44\pm2.79$  \\ \cmidrule{1-1} \cmidrule{3-3} \cmidrule{5-5} \cmidrule{7-7}
jaffe        &  & Centralized &  & $0.68\pm0.02$ & & $38.95\pm1.43$  \\
$\;\;m=676$  &  & Local       &  & $0.64\pm0.01$ & & $42.36\pm1.51$  \\
$\;\;n\;=213$&  & DC(TSVD)    &  & $0.59\pm0.04$ & & $31.46\pm2.46$  \\
             &  & DC(SMOTE)   &  & $0.62\pm0.04$ & & $40.67\pm6.82$  \\ \cmidrule{1-1} \cmidrule{3-3} \cmidrule{5-5} \cmidrule{7-7}
leukemia     &  & Centralized &  & $0.74\pm0.14$ & & $94.64\pm2.95$  \\
$\;\;m=7129$ &  & Local       &  & $0.39\pm0.06$ & & $80.39\pm2.39$  \\
$\;\;n\;=72$ &  & DC(TSVD)    &  & $0.39\pm0.10$ & & $81.61\pm3.94$  \\
             &  & DC(SMOTE)   &  & $0.50\pm0.13$ & & $87.68\pm3.50$  \\ \cmidrule{1-1} \cmidrule{3-3} \cmidrule{5-5} \cmidrule{7-7}
\bottomrule
\end{tabular}
\end{center}
\end{table}
\begin{table}[!t]
  \footnotesize
  \caption{Recognition performance (average $\pm$ standard error) for real-world problems (the second five of ten datasets).}
\label{table:result2}
\begin{center}
\begin{tabular}{lclcccc} 
\toprule
\multicolumn{1}{c}{Dataset} &  & \multicolumn{1}{c}{Method} &  & NMI &  & ACC  \\ \cmidrule{1-1} \cmidrule{1-1} \cmidrule{3-3} \cmidrule{5-5} \cmidrule{7-7}
lung         &  & Centralized &  & $0.69\pm0.05$ & & $88.14\pm2.08$  \\
$\;\;m=3312$ &  & Local       &  & $0.52\pm0.03$ & & $78.06\pm1.50$  \\
$\;\;n\;=203$&  & DC(TSVD)    &  & $0.64\pm0.05$ & & $86.74\pm1.57$  \\
             &  & DC(SMOTE)   &  & $0.65\pm0.03$ & & $85.64\pm2.11$  \\ \cmidrule{1-1} \cmidrule{3-3} \cmidrule{5-5} \cmidrule{7-7}
pixraw10P    &  & Centralized &  & $0.68\pm0.02$ & & $38.00\pm1.79$  \\
$\;\;m=10000$&  & Local       &  & $0.63\pm0.04$ & & $41.00\pm3.12$  \\
$\;\;n=100$  &  & DC(TSVD)    &  & $0.61\pm0.03$ & & $29.00\pm0.89$  \\
             &  & DC(SMOTE)   &  & $0.65\pm0.04$ & & $35.00\pm1.41$  \\ \cmidrule{1-1} \cmidrule{3-3} \cmidrule{5-5} \cmidrule{7-7}
Prostate\_GE &  & Centralized &  & $0.40\pm0.10$ & & $84.09\pm3.63$  \\
$\;\;m=5966$ &  & Local       &  & $0.28\pm0.04$ & & $75.30\pm2.20$  \\
$\;\;n\;=102$&  & DC(TSVD)    &  & $0.31\pm0.07$ & & $78.36\pm2.37$  \\
             &  & DC(SMOTE)   &  & $0.58\pm0.08$ & & $90.18\pm2.01$  \\ \cmidrule{1-1} \cmidrule{3-3} \cmidrule{5-5} \cmidrule{7-7}
TOX-171      &  & Centralized &  & $0.37\pm0.03$ & & $59.70\pm2.62$  \\
$\;\;m=5789$ &  & Local       &  & $0.30\pm0.01$ & & $49.52\pm1.51$  \\
$\;\;n\;=171$&  & DC(TSVD)    &  & $0.34\pm0.04$ & & $49.68\pm3.72$  \\
             &  & DC(SMOTE)   &  & $0.38\pm0.02$ & & $60.24\pm2.12$  \\ \cmidrule{1-1} \cmidrule{3-3} \cmidrule{5-5} \cmidrule{7-7}
warpAR10P    &  & Centralized &  & $0.64\pm0.02$ & & $30.00\pm2.96$  \\
$\;\;m=2400$ &  & Local       &  & $0.59\pm0.02$ & & $34.87\pm1.39$  \\
$\;\;n\;=130$&  & DC(TSVD)    &  & $0.59\pm0.03$ & & $30.77\pm3.61$  \\
             &  & DC(SMOTE)   &  & $0.58\pm0.05$ & & $30.77\pm3.61$  \\ \cmidrule{1-1} \cmidrule{3-3} \cmidrule{5-5} \cmidrule{7-7}
\bottomrule
\end{tabular}
\end{center}
\end{table}
Next, we evaluate the performance of the methods when applied to the binary and multi-class classification problems presented by \cite{lecun1998mnist,samaria2994parameterisation} and feature selection datasets \footnote{available at \url{http://featureselection.asu.edu/datasets.php.}}.
Note that these datasets were used in \cite{imakura2021interpretable}.
\par
We considered the case where each dataset is distributed into six parties: $c=2$ and $d=3$.
The performance of each method was evaluated using a five-fold cross-validation framework.
\par
For the DC analysis, we set the dimensionality of intermediate representations to $\widetilde{m}_{i,j} = 15$ for all parties and the number of anchor data to $r=2500$.
We used the ridge regression for analyzing the collaboration representation (Step 9 in Algorithm 1) and the decision tree for interpretable model (Step 13 in Algorithm~\ref{alg:IDC}).
We set five as the maximum number of branch node splits. 
For the TSVD-based method, we set the rank of TSVD to $20$.
For the proposed SMOTE-based method, we set $(k,\alpha) = (50,1.5)$.
\par
The numerical results for each method are presented in Tables~\ref{table:result} and~\ref{table:result2}  for ten datasets.
Tables~\ref{table:result} and~\ref{table:result2} show that the recognition performance of the proposed method is better than that of {\bf DC(TSVD)} and {\bf Local} on most datasets.
\subsection{Remarks}
In numerical experiments, the proposed {\bf DC(SMOTE)} exhibits a high level of both recognition accuracy and data confidentiality for artificial and real-world datasets.
Numerical experiments demonstrate that the recognition performance of the proposed {\bf DC(SMOTE)} is improved owing the contribution of the extension of SMOTE using a large $\alpha$ and a large $k$.
\section{Conclusions}
This study was motivated by an integrated analysis of multiple financial institutions.
Here, extensive cross-institutional communication can be a significant problem in social implementation.
DC analysis has recently been developed as one of the more logical options for an integrated analysis with small cross-institutional communications.
In this study, we specifically focused on the interpretable DC analysis and proposed a SMOTE-based technique for anchor data construction to improve the accuracy and privacy.
For the proposed method, we extended SMOTE to allow even extrapolation and used a large $k$ of $k$-nearest neighbors.
Numerical results demonstrate the efficiency of the proposed SMOTE-based method over the existing anchor data constructions owing to the contribution of the extension of SMOTE.
Specifically, the proposed method achieves 9 percentage point and 38 percentage point performance improvements regarding accuracy and essential feature selection, respectively, over existing methods for an income dataset.
\par
Privacy-preserving integrated analyses are an essential challenge to address in real-world applications, such as medical, financial, and manufacturing data analyses.
The DC analysis using the proposed SMOTE-based anchor data construction is a breakthrough technology for such types of distributed data analyses.
Additionally, the proposed method provides another use of SMOTE not for imbalanced data classifications but for a key technology of privacy-preserving integrated analysis.
\par
On the other hand, in social implementation, the integrated analysis faces some difficulties, such as loss of data and batch effects.
This study did not consider these difficulties to be inherent in social implementation.
In future, we will intend to address these difficulties.
We also intend to further analyze the confidentiality of the proposed SMOTE-based method and develop software.
\section*{Acknowledgements}
This work was supported in part by the New Energy and Industrial Technology Development Organization (NEDO), Japan Science and Technology Agency (JST) (No. JPMJPF2017), the Japan Society for the Promotion of Science (JSPS), Grants-in-Aid for Scientific Research (Nos. JP19KK0255, JP21H03451, JP22H00895, JP22K19767).
\bibliography{mybibfile}

\begin{thebibliography}{10}
\expandafter\ifx\csname url\endcsname\relax
  \def\url#1{\texttt{#1}}\fi
\expandafter\ifx\csname urlprefix\endcsname\relax\def\urlprefix{URL }\fi

\bibitem{bishop2006pattern}
C.~M. Bishop, Pattern Recognition and Machine Learning (Information Science and
  Statistics), Springer-Verlag Berlin, Heidelberg, 2006.

\bibitem{bunkhumpornpat2009safe}
C.~Bunkhumpornpat, K.~Sinapiromsaran, C.~Lursinsap, {Safe-level-SMOTE:}
  safe-level-synthetic minority over-sampling technique for handling the class
  imbalanced problem, in: Pacific-Asia conference on knowledge discovery and
  data mining, Springer, 2009.

\bibitem{chawla2002smote}
N.~V. Chawla, K.~W. Bowyer, L.~O. Hall, W.~P. Kegelmeyer, {SMOTE:} synthetic
  minority over-sampling technique, Journal of artificial intelligence research
  16 (2002) 321--357.

\bibitem{feng2022vertical}
S.~Feng, Vertical federated learning-based feature selection with
  non-overlapping sample utilization, Expert Systems with Applications 208
  (2022) 118097.

\bibitem{fisher1936use}
R.~A. Fisher, The use of multiple measurements in taxonomic problems, Annals of
  human genetics 7~(2) (1936) 179--188.

\bibitem{han2005borderline}
H.~Han, W.-Y. Wang, B.-H. Mao, {Borderline-SMOTE:} a new over-sampling method
  in imbalanced data sets learning, in: International conference on intelligent
  computing, Springer, 2005.

\bibitem{he2008adasyn}
H.~He, Y.~Bai, E.~A. Garcia, S.~Li, {ADASYN:} adaptive synthetic sampling
  approach for imbalanced learning, in: 2008 IEEE international joint
  conference on neural networks (IEEE world congress on computational
  intelligence), IEEE, 2008.

\bibitem{he2004locality}
X.~He, P.~Niyogi, Locality preserving projections, in: Advances in neural
  information processing systems, 2004.

\bibitem{imakura2021interpretable}
A.~Imakura, H.~Inaba, Y.~Okada, T.~Sakurai, Interpretable collaborative data
  analysis on distributed data, Expert Systems with Applications 177 (2021)
  114891.

\bibitem{imakura2019complex}
A.~Imakura, M.~Matsuda, X.~Ye, T.~Sakurai, Complex moment-based supervised
  eigenmap for dimensionality reduction, in: Proceedings of the AAAI Conference
  on Artificial Intelligence, vol.~33, 2019.

\bibitem{imakura2020data}
A.~Imakura, T.~Sakurai, Data collaboration analysis framework using
  centralization of individual intermediate representations for distributed
  data sets, ASCE-ASME Journal of Risk and Uncertainty in Engineering Systems,
  Part A: Civil Engineering 6 (2020) 04020018.

\bibitem{imakura2021collaborative}
A.~Imakura, X.~Ye, T.~Sakurai, Collaborative data analysis: Non-model
  sharing-type machine learning for distributed data, in: Uehara H., Yamaguchi
  T., Bai Q. (eds) Knowledge Management and Acquisition for Intelligent
  Systems. PKAW 2021. Lecture Notes in Computer Science, vol. 12280, 2021.

\bibitem{imakura2021collaborative2}
A.~Imakura, X.~Ye, T.~Sakurai, Collaborative novelty detection for distributed
  data by a probabilistic method, in: Proceedings of The 13th Asian Conference
  on Machine Learning (ACML 2021), 2021.

\bibitem{jolliffe1986principal}
I.~T. Jolliffe, Principal component analysis and factor analysis, in: Principal
  component analysis, Springer, 1986, pp. 115--128.

\bibitem{konevcny2016federated}
J.~Kone{\v{c}}n{\`y}, H.~B. McMahan, F.~X. Yu, P.~Richtarik, A.~T. Suresh,
  D.~Bacon, Federated learning: Strategies for improving communication
  efficiency, in: NIPS Workshop on Private Multi-Party Machine Learning, 2016.

\bibitem{lecun1998mnist}
Y.~LeCun, The {MNIST} database of handwritten digits, http://yann. lecun.
  com/exdb/mnist/.

\bibitem{lee2000algorithms}
D.~D. Lee, H.~S. Seung, Algorithms for non-negative matrix factorization, in:
  Proceedings of the 13th International Conference on Neural Information
  Processing Systems, MIT Press, 2000.

\bibitem{li2019survey}
Q.~Li, Z.~Wen, Z.~Wu, S.~Hu, N.~Wang, B.~He, A survey on federated learning
  systems: Vision, hype and reality for data privacy and protection, arXiv
  preprint (2019) arXiv:1907.09693.

\bibitem{li2020federated}
T.~Li, A.~K. Sahu, M.~Zaheer, M.~Sanjabi, A.~Talwalkar, V.~Smith, Federated
  optimization in heterogeneous networks, Proceedings of Machine Learning and
  Systems 2 (2020) 429--450.

\bibitem{li2017locality}
X.~Li, M.~Chen, F.~Nie, Q.~Wang, Locality adaptive discriminant analysis, in:
  Proceedings of the 26th International Joint Conference on Artificial
  Intelligence, AAAI Press, 2017.

\bibitem{mcmahan2016communication}
H.~B. McMahan, E.~Moore, D.~Ramage, S.~Hampson, et~al., Communication-efficient
  learning of deep networks from decentralized data, arXiv preprint (2016)
  arXiv:1602.05629.

\bibitem{mizoguchi2022application}
A.~Mizoguchi, A.~Imakura, T.~Sakurai, Application of data collaboration
  analysis to distributed data with misaligned features, Informatics in
  Medicine Unlocked 32 (2022) 101013.

\bibitem{ni2022federated}
X.~Ni, X.~Shen, H.~Zhao, Federated optimization via knowledge codistillation,
  Expert Systems with Applications 191 (2022) 116310.

\bibitem{samaria2994parameterisation}
F.~Samaria, A.~Harter, Parameterisation of a stochastic model for human face
  identification, in: Proceeding of IEEE Workshop on Applications of Computer
  Vision, 1994.

\bibitem{strehl2002cluster}
A.~Strehl, J.~Ghosh, Cluster ensembles---a knowledge reuse framework for
  combining multiple partitions, Journal of machine learning research 3~(Dec)
  (2002) 583--617.

\bibitem{sugiyama2007dimensionality}
M.~Sugiyama, Dimensionality reduction of multimodal labeled data by local
  {Fisher} discriminant analysis, Journal of machine learning research 8~(May)
  (2007) 1027--1061.

\bibitem{yang2019federated}
Q.~Yang, Y.~Liu, T.~Chen, Y.~Tong, Federated machine learning: Concept and
  applications, ACM Transactions on Intelligent Systems and Technology 10~(2)
  (2019) Article 12.

\bibitem{ye2019distributed}
X.~Ye, H.~Li, A.~Imakura, T.~Sakurai, Distributed collaborative feature
  selection based on intermediate representation, in: The 28th International
  Joint Conference on Artificial Intelligence (IJCAI-19), 2019.

\end{thebibliography}
\bibliographystyle{elsart-num-sort}
\end{document}